\documentclass[11pt, a4paper]{article}

\usepackage{amsmath,amssymb}
\usepackage{graphicx}
\usepackage{xspace}
\usepackage{algorithm2e}
\usepackage{comment}
\usepackage[font={small}]{caption}
\newcommand{\url}[1]{{\tt\scriptsize #1}}

\newcommand{\PISq}{PI\ensuremath{^2}\xspace}
\newcommand{\PIBB}{\mbox{PI$^{\mbox{\sf\tiny BB}}$}\xspace}
\newcommand{\CMAES}{\textsc{CMA-ES}\xspace}

\newcommand{\equ}[1]{(\ref{equ_#1})}

\newcommand{\mymath}[1]{\ensuremath{#1}\xspace}
\newcommand{\mymathbf}[1]{\mymath{\mathbf{#1}}}
\newcommand { \mylabel}[1]{{\mbox{\tiny{\sf #1}}}}

\newcommand { \vq  } {\mymathbf{  q      }}
\newcommand { \vx  } {\mymathbf{  x      }}

\newcommand { \vp  } {\mymathbf{  \theta }}
\newcommand { \vpopt  } {\mymath{\vp^{*}}}
\newcommand { \vg  } {\mymathbf{  g }}
\newcommand { \mSigma  } {\mymathbf{  \Sigma }}

\newcommand{\N}[2]{\mathcal{N}(#1,#2)}

\newcommand{\vpinit}{\mymath{\vp^\mylabel{init}}}
\newcommand{\vpnew}{\mymath{\vp^\mylabel{new}}}
\newcommand{\lambdainit}{\mymath{\lambda^\mylabel{init}}}
\newcommand{\lambdamin} {\mymath{\lambda^\mylabel{min}}}

\usepackage{ifthen}
\newboolean{anonymous}
\setboolean{anonymous}{false}

\usepackage{apacite}
\usepackage[]{authblk}
\usepackage[margin=1.in]{geometry}
\graphicspath{{./images-bin/}}

\date{}

\begin{document}

\author[1,2,3]{Freek Stulp}
\author[1,2]{Pierre-Yves Oudeyer \footnote{Corresponding author: Pierre-Yves Oudeyer (Postal address: Inria, 200, avenue de la vieille tour, 33405 Talence, France; email: pierre-yves.oudeyer@inria.fr)}}

\affil[1]{Inria, France}
\affil[2]{ENSTA ParisTech, Universit\'e Paris-Saclay, France}
\affil[3]{German Aerospace Center (DLR), Germany}
\title{Proximodistal Exploration in Motor Learning as an Emergent Property of Optimization}
\maketitle

\abstract{To harness the complexity of their high-dimensional bodies during sensorimotor development, infants are guided by patterns of freezing and freeing of degrees of freedom. For instance, when learning to reach, infants free the degrees of freedom in their arm proximodistally, i.e. from joints that are closer to the body to those that are more distant. Here, we formulate and study computationally the hypothesis that such patterns can emerge spontaneously as the result of a family of stochastic optimization processes (evolution strategies with covariance-matrix adaptation), without an innate encoding of a maturational schedule.  In particular, we present simulated experiments with an arm where a computational learner progressively acquires reaching skills through adaptive exploration, and we show that a proximodistal organization appears spontaneously, which we denote PDFF (ProximoDistal Freezing and Freeing of degrees of freedom). We also compare this emergent organization between different arm morphologies -- from human-like to quite unnatural ones -- to study the effect of different kinematic structures on the emergence of PDFF. }
\\

{\bf Research highlights.}
\begin{itemize}
  \item We propose a general, domain-independent hypothesis for the developmental organization of freezing and freeing of degrees of freedom observed both in infant development and adult skill acquisition, such as proximo-distal exploration in learning to reach.
  \item We introduces a computational model based on basic principles of stochastic optimization, and show how proximodistal freezing and freeing of degrees of freedom arises as an emergent property of this model.
  \item We analyze the influence of human arm structure on the patterns of freezing and freeing of degrees of freedom in simulated reaching tasks.
\end{itemize}

\section{Introduction}

As Bernstein emphasized \cite{Bernstein67}, a great mystery in infant motor development is to understand how they can learn motor skills efficiently given a a complex non-linear body with many degrees of freedoms. Robots face the same problems, and this issue has similarly been the object of many studies in the recent years \cite{vijayakumar2005incremental,baranes11interaction,kober11policy,baranes:hal-00788440,stulp13robot}. 
Learning motor skills involves experimenting with one's own body under limited time resources, and thus only a small fraction of physically possible movements can be sampled within the first years of life. Thus, as argued for example in \cite{Berthier99} and theoretically analyzed from a machine learning perspective \cite{oudeyer:hal-00788611}, learning strategies based on simple forms of trial and error cannot lead to efficient learning in such contexts.

Several strands of research have studied families of mechanisms that could constrain and guide motor learning processes. In particular, Bernstein established a motor development perspective based on staged learning processes where some degrees of freedoms were first frozen, transforming a complex learning problem in a simpler one, and then progressively freed, allowing the learner to take advantage of the full potential of its body \cite{Bernstein67}. A number of experimental studies allowed to confirm this perspective. For example, Berthier et al. \cite{Berthier99} showed that the development of early reaching in infants \cite{Bertenthal98} followed a proximodistal structure, where infants first learnt to reach by freezing the elbow and the hand, while varying shoulder and trunk movements, and then progressively used more distal joints of the elbow and hand. Studies in adult motor skill acquisition showed similar patterning of freezing and freeing of degrees of freedom, applied to the acquisition of racket skills \cite{southard1987changing}, soccer \cite{hodges2005changes} or skiing \cite{vereijken1992free}. Other experimental observations have shown the complexity and context-dependance of this form of patterning, where for example infant reaching with different postural constraints could lead to higher use of elbow with respect to the shoulder \cite{thelen1993transition}. 

Several hypotheses explaining the underlying mechanisms leading to such staged motor learning schedules were formulated so far. For example, Berthier et al.\cite{Berthier99} suggested that these learning schedules could be innate and due to the progressive neuromuscular development, where physiological maturation of motor neurons along the corticospinal tract \cite{kuypers1981anatomy, jansen1990perinatal} could potentially lead to an initial limitation in the control of distal degrees of freedom. Yet, the extent to which physiological maturation can constrain motor exploration is still unclear in the infant \cite{adolph2005physical}, and does not provide an explanation of the underlying mechanisms which drive freezing and freeing of degrees of freedom in adult motor learning. 

In this article, we formulate, explore and analyze another (possibly complementary) hypothesis from a computational modelling perspective. This hypothesis is formulated within the optimal control framework of motor learning, where the learner uses exploration to find a motor program which minimizes a given cost (or maximizes an objective function) ~\cite{todorov04optimality, berthier2005approximate}. The hypothesis we study states that staged learning schedules with freezing and progressive freeing of degrees of freedom can self-organize spontaneously as a result of the interaction between certain families of stochastic optimization methods (which drive exploration of the learner) with physical properties of the body, and without involving physiological maturation. In particular, we present simulated experiments with a 6-DOF arm where a computational learner progressively acquires reaching skills (i.e. minimizing a cost to reach), and we show that a proximodistal organization appears spontaneously, which we denote PDFF (Proximo Distal Freezing and Freeing of degrees of freedom). We also compare the emergent structuration as different arm structures are used -- from human-like to quite unnatural ones -- to study the effect of different kinematic structures on the emergence of PDFF. 

In these experiments\footnote{The Matlab code used to generate and visualize the results in this article is available as open source, and can be downloaded here: %
\ifthenelse{\boolean{anonymous}}{
\url{http://}URL anonymized
}{
{\tt\tiny https://github.com/stulp/dmp\_bbo/archive/proximodistalmaturation.zip}
}
}, the reaching task is learned by applying stochastic optimization to optimize the parameters of a movement policy.
The algorithm we use -- \PIBB, a special case of \PISq -- is based on covariance matrix adaptation through weighted averaging, which is a concept present in a wide range of optimization frameworks~\cite{arnold11informationgeometric,rubinstein04crossentropy,stulp13robot}. 
Covariance matrix adaptation allows the algorithm to determine dynamically the appropriate exploration magnitude and direction for each joint in order to progress fastest towards the goal at any given point in the development. In the context of PDFF, increasing and decreasing the exploration corresponds to \emph{freeing} and \emph{freezing} joints respectively. 

However, to our knowledge, methods for exploration through covariance matrix adaptation were so far analyzed only from an engineering perspective and in terms of speed to find optimal controllers. Here, on the contrary, we use these general methods as tools to modeling processes of exploration during motor learning in infants and study the patterns of freeing and freezing of DOFs that they generate. Preliminary work in this direction was presented in \cite{stulp12emergent, stulp13adaptive}, but was based on more complex and specific optimization algorithms, did not include detailed analysis of results, and did not study how different morphologies of the body impacted the resulting patterns of exploration. 

Here, we use a simple and generic form of covariance matrix adaptation -- \PIBB -- and study how it spontaneously generates PDFF exploration patterns in the context of several arm morphologies. In the two analysis sections, we further study these results by considering the effect of joints on the cost in a static context. We first perform a sensitivity analysis by quantifying the effect of perturbing individual joint on the main cost component. We then analyze the interactions between joints by determining the effects of perturbing distal joints in the context of perturbations to proximal joints. The results of this analysis provide a deeper understanding of \emph{why} PDFF arises during the stochastic optimization.

\section{Limitations of Prior Research}

Many computational models have studied how prewired stages or patterns of freezing and freeing of degrees of freedom could contribute or hinder learning of motor skills in high-dimensions. Some studies considered the impact of alternation of freezing and freeing phases upon robot learning of swinging skills ~\cite{Berthouze04}, studied how the pace of the sequencing of discrete stages~\cite{Bongard10, Grupen03,  Lee07b} or of the continuous increase of explored values of DOFs along a proximodistal scheme~\cite{baranes11interaction} could be adaptively and non-linearly controlled by learning progress and lead to efficient motor learning in high-dimensional robots. Other related models have explored how the progressive freeing of degrees of freedom in the perceptual space \cite{Nagai06, French02}, in the environment \cite{Asada98}, or in the structure of neural networks for learning abstractions \cite{Elman93, westermann2007neuroconstructivism} could guide the acquisition of sensorimotor and cognitive skills.

In all these models, the global scheduling of freezing and freeing degrees of freedom is encoded by the engineer (but the rhythm of progression from stage to stage can be adaptive as in~\cite{baranes11interaction,Lee07b}). Some models have explored explicitly the evolutionary mechanisms that could generate and select such innate maturational schedules \cite{Cangelosi99, Matos07}. 

A related model is presented in Schlesinger et al.~\cite{Schlesinger00}. It is most similar to ours in that it also uses a kinematically simulated arm, and explores how evolutionary-like stochastic optimization methods  can lead ``several constraints to appear to fall out as a consequence of a relatively simple trial-and-error learning algorithm''~\cite{Schlesinger00}, one of them being the locking of joints. Movement policies are represented as four-layer feedforward neural networks, which are trained through evolutionary learning. In terms of the experimental setup, one main difference to our work is that we consider higher-dimensional systems -- 10 DOF instead of 3DOF -- and use one learning agent instead of a population of 100. A second difference is that we employ a different family of stochastic optimization techniques, which has more flexibility in that it can dynamically update ranges of exploration during single agent learning. While the model in Schlesinger et al.~\cite{Schlesinger00} only accounted for the freezing of some degrees of freedom as a result of optimization, the flexibility of our learning model allows us to find the entire developmental pattern outlined by Bernstein~\cite{Bernstein67}: freezing of degrees of freedom followed by progressive and ordered freeing of degrees of freedom. We also consider various arm morphologies to show how the emergent scheduling adapts to the peculiarities of a given kinematic structure.

\section{Learning to Reach with Stochastic Optimization}

The methods, results and discussions of the experiments are distributed over three sections, corresponding to the three experiments conducted. In this first section, we describe an experiment in which a parameterized policy generates reaching movements, where the parameters of the policy are optimized through stochastic optimization.

\subsection{Methods}

\subsubsection{Arm Model}

\begin{figure}[ht]
  \centering
  \includegraphics[width=1.0\columnwidth]{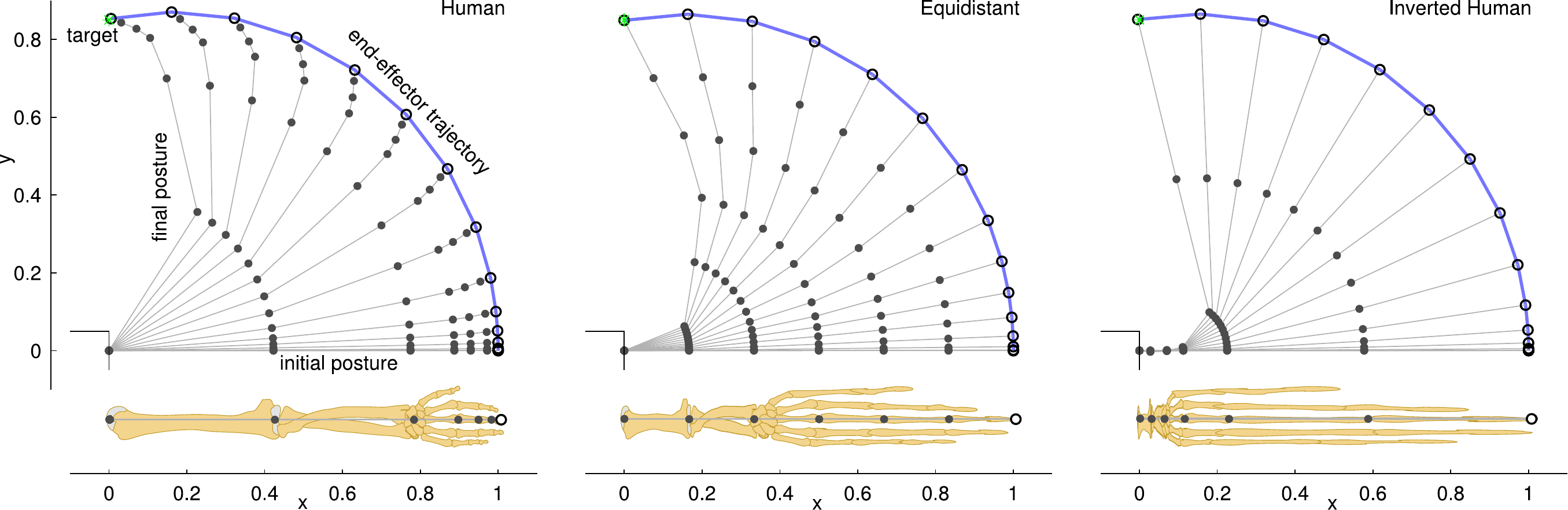}
  \caption{\label{fig_task} Visualization of the task, which is to reach for a specific target location, in this case $[0,0.85]$. The arm starts horizontally, and the movement to the target is visualized with a stroboscopic snapshots.  
  }
\end{figure}

The evaluation task in this paper consists of a kinematically simulated arm with $M=6$ degrees of freedom, and a normalized length of 1. To study the effect of different kinematic structures on maturation, we use three sets of relative link lengths, depicted at the bottom of \figurename~\ref{fig_task}: 1)~typical relative link lengths of a human arm; 2)~equidistant link lengths; 3)~`inverted' human arm, i.e. with short link lengths first.

\subsubsection{Task Specification}

The main aim of the task is to reach for a specific target with a 0.5s movement, visualized in \figurename~\ref{fig_task}. The angles and angular velocities of all joints are initially 0, which corresponds to a completely stretched `horizontal' arm.
The cost function in equations \equ{cost_term}-\equ{cost_immediate} consists of three parts, expressing different criteria to be optimized during the learning process:

\begin{description}
  \item[Terminal cost $||\vx_{t_N}-\vx^g||^2$.]
  The distance between the 2-D Cartesian coordinates of the end-effector ($\vx_{t_N}$) at the end of the movement at $t_N$, and the goal $\vx^{g}$. 
  This expresses that we want to reach to the target $\vx^g$ as closely as possible.
  
  \item[Terminal cost $\mbox{\sf max}(\vq_{t_N})$.]
  A cost that corresponds to the largest angle over all the joints at the end of the movement.
  This expresses an end-state comfort effect~\cite{cohen04where}.

  \item[Immediate cost $r_t$]
  The immediate costs at each time step $r_t$ in \equ{cost_immediate} penalize joint accelerations. The weighting term $(M+1-m)$ penalizes DOFs closer to the origin, the underlying motivation being that wrist movements are less costly than shoulder movements for humans, cf.~\cite{theodorou10generalized}\footnote{This cost term was taken from~\cite{theodorou10generalized}. In the context of this paper, it cannot be the reason for the PDFF we shall see in later sections. 
  Rather than favoring PDFF, this cost term actually works \emph{against} it, as proximal joints are penalized \emph{more} for the accelerations that arise due to exploration.}.   
  
\end{description}

\begin{scriptsize}
\begin{align}
\phi_{t_N} &= 10^2||\vx_{t_N}-\vx^g||^2 + \mbox{max}(\vq_{t_N}) & \mbox{Terminal cost}\label{equ_cost_term}\\
r_t &= 10^{-5} \frac{\sum_{m=1}^M(M+1-m)(\ddot{q}_{t,m})^2}{\sum_{m=1}^M (M+1-m)} & \mbox{Immediate cost}\label{equ_cost_immediate}
\end{align}
\end{scriptsize}

The factors $10^2$ and $10^{-5}$ have two purposes: 1)~a scaling factor to compensate for different range of values the different cost components have. 2)~a weighting factor enabling the prioritization of tasks. The order of priorities is: reach close to the target, achieve end-state comfort, minimize accelerations.

\begin{figure}[ht]
  \centering
  \includegraphics[width=1.0\columnwidth]{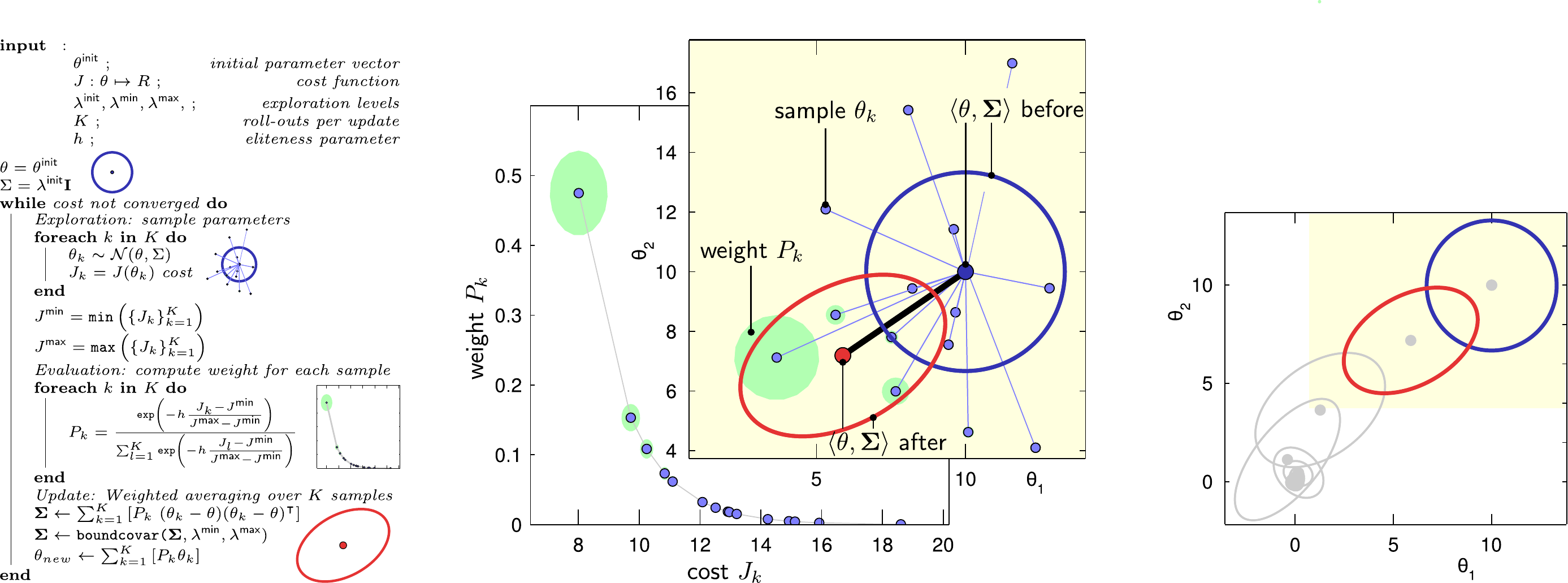}
  \caption{\label{fig_update_visualization} Explanation and visualization of the \PIBB algorithm, using a 2D search space. For illustratory purposes, the cost of a sample \vp is simply the distance to the origin: $J(\vp) = ||\vp||$. Left: \PIBB pseudo-code. Center: Visualization of \emph{one} parameter update with \PIBB. Right: Evolution of the parameters over \emph{several} updates, demonstrating how the distribution converges towards the minimum $\vpopt=[0,0]$.\\
  The algorithm is initialized by setting the mean and covariance parameters $\langle\vp,\mSigma\rangle$  to \vpinit and $\lambdainit\mathbf{I}$ respectively, visualized as a dark blue dot and circle.
  After this initialization, \PIBB then iteratively updates these parameters with the following steps:
  1)~{\em Explore.} Sample $K$ parameter vectors $\vp_k$ from $\N{\vp}{\mSigma}$, and determine the cost $J_k$ of each sample. In the visualization of our illustratory example task $K=15$, and the cost $J(\vp)$ is the distance to the origin $||\vp||$, which lies approximately between 8 and 19 in this example.
  2)~{\em Evaluate.} Determine the weight (probability) $P_k$ of each sample, given its cost. Essentially, low-cost samples have higher weights, and vice versa. The normalized exponentiation function that maps costs to weights is taken directly from the \PISq algorithm, and is visualized in the center graph. Larger green circles correspond to higher weights.
  3)~{\em Update.} Updating the parameters $\langle\vp,\mSigma\rangle$ with weighted averaging. In the visualization, the updated parameters are depicted in red. Because low-cost samples (e.g. a cost of 8-10) have higher weights, they contribute more to the update, and $\vp$ therefore moves in the direction of the optimum $\vpopt=[0,0]$.
  }
\end{figure}

\subsubsection{Policy Representation}

The policy representation encodes how movements are generated, by specifying the acceleration profiles of each joint. It is represented as a linear combination of normalized Gaussian basis functions. The acceleration $\ddot{q}_{m,t}$ of the $m^{\mbox{\scriptsize th}}$ joint at time $t$ is determined as in \equ{policy}, where the parameter vector $\vp_m$ represents the weights of joint $m$.

Intuitively, different basis functions are active at different times during the movement. The first basis function $\Psi_{b=0}(t)$ is most active at the beginning of the movement, and the last $\Psi_{b=B}(t)$ at the end of the movement, with a cascade of basis functions in between. Setting different weights in the parameter vector $\vp$ thus leads to different acceleration profiles during the movement. If $\vp=\mathbf{0}$, then there is no acceleration, and thus no movement.

\begin{small}
\begin{align}
\ddot{q}_{m,t} &= \vg_t^\intercal\vp_m  & \mbox{Acceleration of joint $m$} \label{equ_policy}\\
{[\vg_t]}_b &=\frac{\Psi_b(t)}{\sum_{b=1}^B \Psi_b(t)} & \mbox{Time-dependent basis functions} \label{equ_basisfunctions}\\
\Psi_b(t) &= \exp \left(-(t-c_b)^2/w^2 \right)  & \mbox{Kernel} \label{equ_kernel}
\end{align}
\end{small}

The centers $c_{b=1\dots B}$ of the kernels $\Psi$ are spaced equidistantly in the 0.5s duration of the movement, and all have a width of $w=0.05s$. The number of kernels per joint is $B=5$. Since we do not simulate arm dynamics, the joint velocities and angles are acquired by integrating the accelerations. The end-effector ``hand'' position \vx~is computed with the forward kinematics of the arm.

\subsubsection{Policy Improvement through Stochastic Optimization}

Stochastic optimization is based on iteratively exploring and updating parameters in a search space \vp (\vp is the vector of parameters of a movement policy). 
At each iteration, stochastic optimization algorithms generate $K$ perturbations of the parameter vector $\{\vp_k = \vp+\epsilon_k\}_{k=1}^K$, compute the cost $J_k$ for each perturbation, and update the parameters $\vp \rightarrow \vpnew$ based on these costs. 
This process continues until the costs have converged, or some termination condition is reached.
In this article, the parameter space \vp corresponds to the parameters of the policy. 
Optimizing policy parameters is known as \emph{direct policy search}.



The specific stochastic optimization algorithm we use is \PIBB, short for ``Policy Improvement with Black-Box optimization''~\cite{stulp12policy_hal}. The \PIBB algorithm is explained and visualized in \figurename~\ref{fig_update_visualization}. We recommend readers to consider \figurename~\ref{fig_update_visualization} in detail, as it is important to understanding the rest of this paper. The main equations from \figurename~\ref{fig_update_visualization} are repeated in (\ref{equ_explore})-(\ref{equ_update_mean}).

\begin{footnotesize}
\begin{align}
&\{\vp_k = \vp+\epsilon_k\}_{k=1}^K & \mbox{Explore by sampling $K$ exploratory parameter vectors $\N{\vp}{\mSigma}$}\label{equ_explore}\\
& J_k = J(\vp), P_k = f(J_k) & \mbox{Evaluate by computing $K$ costs and map costs to weights}\\
& \mSigma^\mylabel{new} = \sum_{k=1}^K[P_k (\vp_k-\vp)(\vp_k-\vp)^\intercal]&  \mbox{Update $\mSigma$ for future exploration with weighted averaging.}\label{equ_update_covar}\\
& \vpnew = \sum_{k=1}^K[P_k \vp_k] &  \mbox{Update the mean for future sampling with weighted averaging.}\label{equ_update_mean}
\end{align}
\end{footnotesize}

The core underlying principle in \PIBB relevant to our experiments is using weighted averaging to update the mean \vp (\ref{equ_update_mean}) and covariance matrix \mSigma (\ref{equ_update_covar}) of the sampling distribution. It shares this principle with many other \textit{evolution strategies} algorithms such as  ``Cross-Entropy Methods''~\cite{rubinstein04crossentropy}, ``Covariance Matrix Adaptation -- Evolutionary Strategies''~\cite{hansen01completely} (\CMAES) and ``Policy Improvement with Path Integrals''~\cite{theodorou10generalized} (\PISq). Hoffman et al. have shown that weighted averaging better explains human motion learning than gradient-based algorithms~\cite{hoffmann08optimization}. 

The intuitive meaning is that the perturbations that are sampled at each iteration of learning are themselves adaptive, fostering exploration in directions where the cost decreases fastest. 
Indeed, \PIBB is a special case of both \PISq (without temporal averaging and with covariance matrix adaptation) and \CMAES (without evolution paths), which are state-of-the-art in direct policy search and black-box optimization respectively. 

Here we select \PIBB over \PISq because it is the simplest such algorithm that implements the principle of weighted averaging to update policy parameters; we are interested in studying the formation of staged patterns of freezing and freeing of degrees of freedom (PDFF), not in achieving for instance the fastest possible convergence. For a full explanation of \PIBB, and its relationship to \PISq and \CMAES, we refer to~\cite{stulp12policy_hal}.

\paragraph{Adaptive Exploration through Covariance Matrix Adaptation}

\begin{figure}[ht]
  \centering
  \includegraphics[width=0.6\columnwidth]{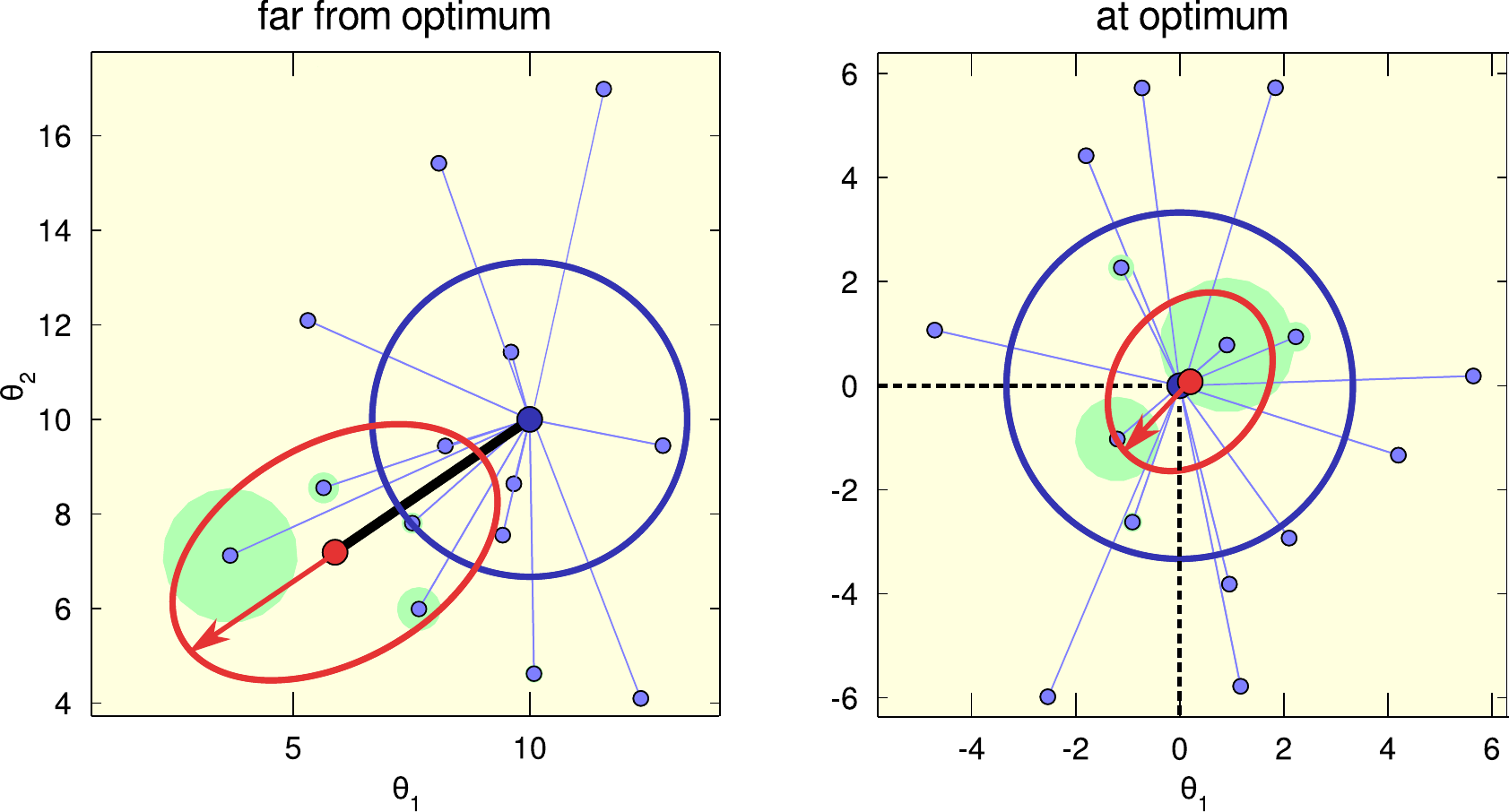}
  \caption{\label{fig_update_begin_end} 
  Left: when the current parameters are \emph{far} from the optimum ($\vpopt=[0,0]$ in this example), the covariance matrix is updated (red ellipse) so that it tends to elongate in the direction of steepest descent, leading to increased exploration along this direction (see arrow).
  Right: when the current parameters are exactly at the optimum, the covariance matrix tends to shrink, leading to decreased exploration.}
\end{figure}

Covariance matrix adaptation allows \PIBB to automatically adapt the exploration so as to generate more samples in the direction of the minimum. Because this property is important for PDFF, we highlight and illustrate it with the two simple examples in \figurename~\ref{fig_update_begin_end}. 

In the left graph in \figurename~\ref{fig_update_begin_end} (repeated from \figurename~\ref{fig_update_visualization}), the current parameters $\vp=[10,10]$ are far from the optimum $\vpopt=[0,0]$. The samples that are closer to \vpopt (to the lower left in the graph) have larger weights (visualized as green circles) than those that are further away from \vpopt. Because the new parameters are weighted with these weights when averaging, the mean \vp moves in the direction of these high-weight, low-cost samples when updating, bringing \vp closer to \vpopt. The same principle applies to the covariance matrix, which becomes elongated -- its largest eigenvalue $\lambda$ increases, and corresponds to the eigenvector pointing more in the direction of \vpopt (visualized as an arrow). More formally, it follows the natural gradient of the cost with respect to the parameters~\cite{arnold11informationgeometric}. The effect is that exploration increases, and in the direction of low-cost regions in the parameter space. 

In the right graph, the initial covariance matrix is the same, but the current parameters $\vp=[0,0]$ are now perfectly at the optimum $\vpopt=[0,0]$. For this reason, samples closer to \vp will have a lower cost $J(\vp) = ||\vp||$, and thus a higher weight. Note the larger green circles are all close to the center. Therefore, \vp hardly moves after updating; this is desirable, because \vp is already at the optimum. However, the covariance matrix shrinks (see smaller eigenvalue as arrow) because closer samples have higher weights. Therefore, exploration decreases in all directions.

This adaptive exploration behavior may also be observed in the right graph in \figurename~\ref{fig_update_visualization}; in the first few updates the covariance matrix elongates towards the optimum $\vpopt=[0,0]$, but once it is reached, the covariance matrix shrinks and exploration decreases. This behavior is especially apparent in this idealized example, which uses only a 2-D search space and a very benign cost function; however, the general principle also applies to high-dimensional spaces and discontinuous cost functions\ifthenelse{\boolean{anonymous}}{.}{, as demonstrated in~\cite{stulp12adaptive}.} 

\subsubsection{Application of \PIBB to the Reaching Task}

In this article, we apply \PIBB to the parameters of the policy representation previously described. 
Each joint has its own parameters $\vp_m$ and covariance matrix $\mSigma_m$, which are iteratively updated with \PIBB. The input parameters of \PIBB are set as follows. The initial parameter vector is $\vpinit=\mathbf{0}$, which means the arm is completely stretched, and not moving at all over time. 
The initial and minimum exploration magnitude of each joint $m$ is set to $\lambdainit=\lambdamin=0.05$. 
The number of trials per update is $K=20$, and the eliteness parameter is $h=10$, the default value suggested by~\cite{theodorou10generalized}\footnote{High values of the eliteness $h$ lead only a few samples to contribute to the weighted averaging. $h=0$ would give all samples equal weight independent of the cost, and no learning would occur. As \cite{hansen01completely} note: ``In general, the selection related parameters [such as $h$] are comparatively uncritical and can be chosen in a wide range without disturbing the adaptation procedure.'' In fact, the same parameter settings have been used in entirely different domains, for instance to optimize robot control policies~\cite{theodorou10generalized}.} A separate stochastic optimization session was run 10 times for each of the 20 target points, i.e. 200 sessions per arm structure.  

The exploration magnitude $\lambda_m$ of a particular joint $m$ at some point during the learning process is defined as the maximum eigenvalue of the covariance matrix $\mSigma_m$. Initially, $\lambda_m$ is $\lambdainit$, because the all eigenvalues of the initial diagonal $\mSigma=\lambdainit\mathbf{I}$ are $\lambdainit$. The length of the two arrows in \figurename~\ref{fig_update_begin_end} visualize $\lambda_m$ for non-diagonal covariance matrices, which may arise as \mSigma is updated.

In the context of this paper, we consider joints that have low exploration magnitudes $\lambda_m$ to be `frozen', whereas those with high $\lambda_m$ are `free'. 

\subsection{Results}

\begin{figure}[ht]
  \centering
  \includegraphics[width=0.6\columnwidth]{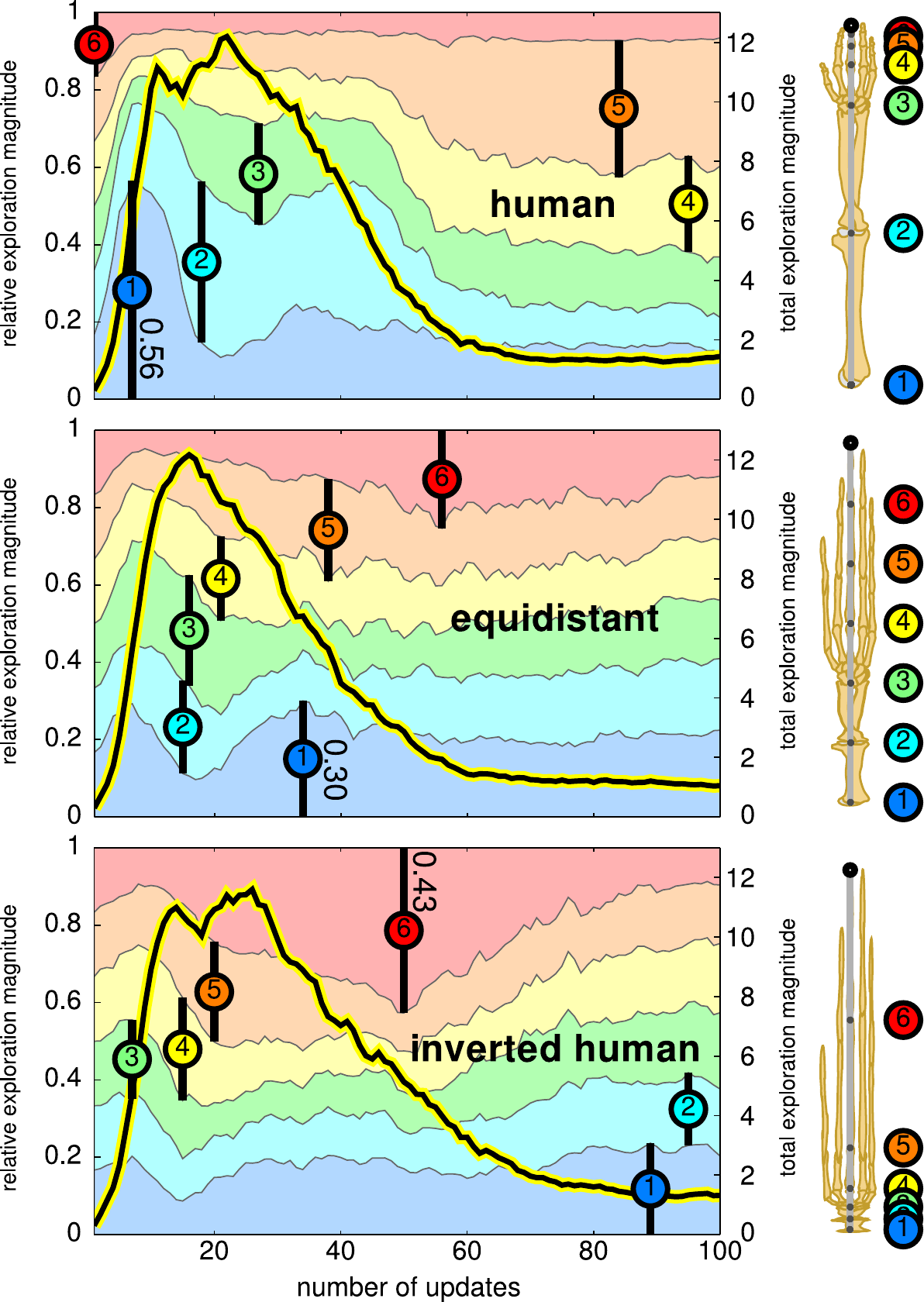}
  \caption{\label{fig_results} Results of stochastic optimization for the three arm structures.The thick black/yellow line represents the total exploration magnitude (right $y$-axis) and relative exploration magnitudes of the 6 individual joints as learning progresses (left $y$-axis).}
\end{figure}

\figurename~\ref{fig_results} presents the results of applying \PIBB to the three different arm structures. Each graph plots the total exploration magnitude over all joints $\sum_{m=1}^M \lambda_m$ (thick yellow/black line) and the relative exploration magnitude $\lambda_m/\sum_{m=1}^M \lambda_m$ per joint (colored patches) as learning progresses over 100 updates, which corresponds to $2000=100\cdot K$ roll-outs. All values are averaged over the 20 target points and 10 optimization sessions per target point. The thick vertical bars indicates when a joint reached its maximum relative exploration magnitude (position on $x$-axis), as well as the magnitude itself (the height of the bar). For example, for the human arm structure (top graph), the first joint achieves a maximum relative exploration magnitude of 0.56 at update 7. This means that 56\% of the exploration is accounted for by only the most proximal (first) joint. 

\begin{figure}[ht]
  \centering
  \includegraphics[width=0.5\columnwidth]{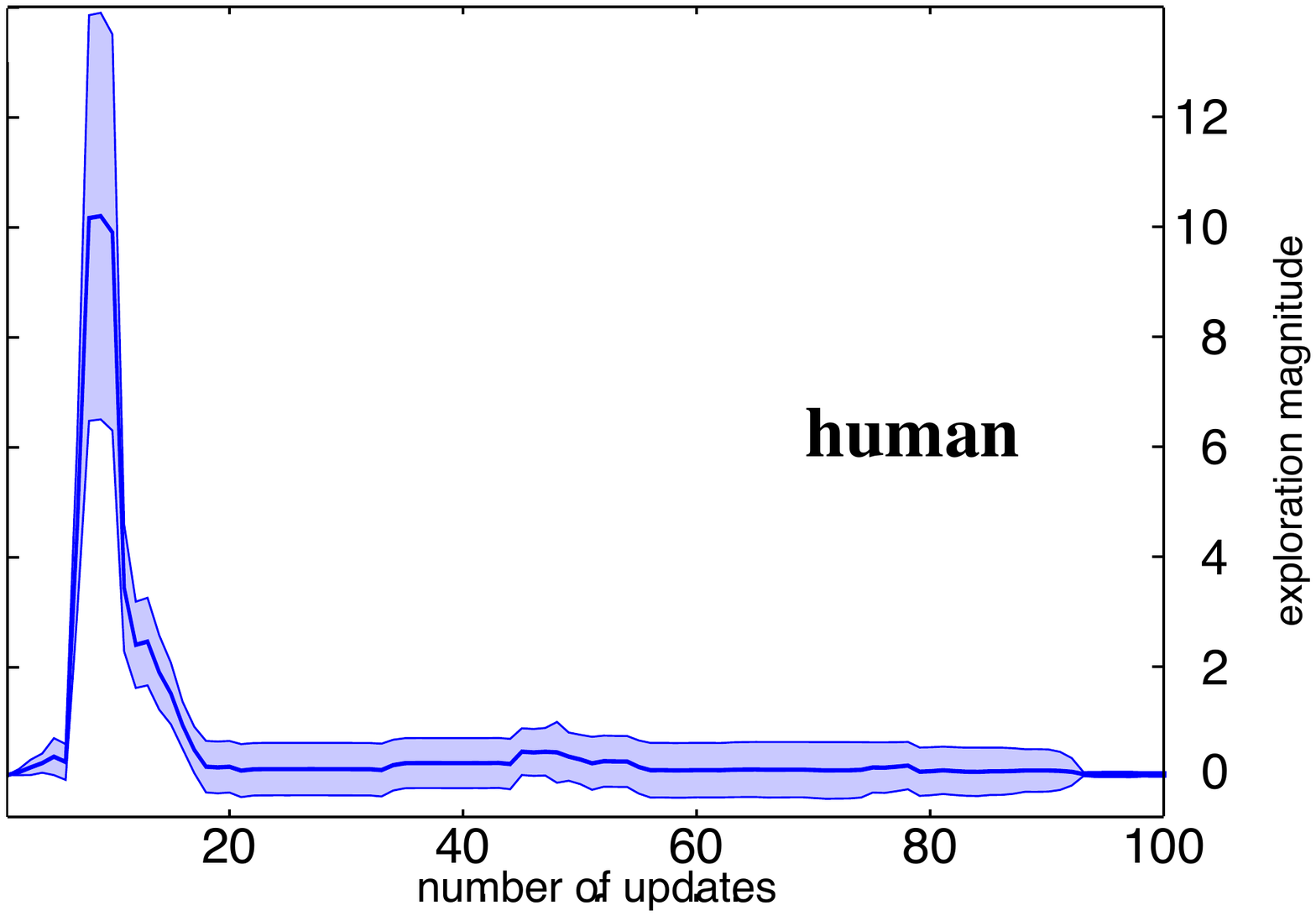}
  \caption{\label{fig_results_variance} Mean and standard deviation in the exploration magnitude of the first joint in the human arm at each update, after aligning the 200 sessions for this arm with dynamic time warping. This figure illustrates that the variance is higher when exploration is highest, as is to be expected.
  The variances for the other joints, not plotted here, are lower.}
\end{figure}

To investigate the robustness of the method against initial conditions, \figurename~\ref{fig_results_variance} plots the variability around the mean of the exploration in the first joint in the human arm for the 200 learning sessions for this arm type. In generating this figure, we noticed that the exact onset of increasing exploration (freeing degrees of freedom) is influenced by the stochastic nature of the optimization algorithm. To factor this out, we have applied dynamic time warping~\cite{sakoe1978dynamic} to the exploration curves of the 200 individual learning sessions before computing the mean and variance.

\subsection{Discussion}

For all arm structures, we see that the total exploration (thick black/yellow line) initially increases, indicating that DOFs are globally freed. After achieving a maximum total exploration at around update 20-25, exploration then decreases again once the task has been learned. This behavior is a direct consequence of the adaptive exploration described in 
\figurename~\ref{fig_update_begin_end}\ifthenelse{\boolean{anonymous}}{.}{, and is also observed in~\cite{stulp12emergent,stulp12adaptive}.}

The relative exploration magnitude between the joints, however, shows quite a different development for the different arm structures. For the human arm (top graph), the most proximal joint is already responsible for more than 50\% of the exploration (0.56) after 7 updates. This joint has been freed, whereas the others are frozen; the three distal joints account for less than 20\% of exploration at update 7. 
For the human arm, we see that joints 1, 2, 3 achieve their maximum relative exploration of 0.56, 0.42 and 0.26 at updates 7, 18 and 27 respectively. In conclusion, we clearly see that the first three joints are freed in a proximodistal order\footnote{Note that the 6th joint achieves its maximum at update 1. This is not because it is freed very early, but rather because it is almost frozen throughout the learning process, and thus achieves its maximum when the exploration is initially the same for all joints}.

For the equidistant arm, the order in which the joints achieve their maxima is 2,3,4,1,5,6. Thus, apart from the most proximal joint, we again see a proximodistal freeing of joints. When considering the maximum relative magnitudes of the exploration (vertical bars), we see however that the freeing/freezing of joints is much less less pronounced. None of the maxima exceeds 0.3, which is in contrast with the human arm, where the most proximal joint is responsible for 0.56 if the exploration. For the equidistant arm, the exploration is thus spread out over the joints much more, rather than being focused in only one or several joints. 

Finally, for the inverted human arm, the order in which joints are freed is 3,4,5,6,1,2. This time, the bulk of the exploration is being done by joint 6, which accounts for more than 42\% of exploration at update 50. The other joints again never exceed 0.3. Thus, for this rather unnatural arm, we see more exploration in the distal joints; only later on do the proximal joints 1 and 2 achieve their maximum values. 

In summary, the results show that there is a consistent emergent organization of exploration over time in all arm structures. The PDFF organization is quite pronounced in the human arm, where the order of freeing is 1,2,3,5,4, and the exploration switches most clearly (i.e. high relative magnitudes of exploration) from one joint to the other. It is important to realize that this effect emerges solely from adaptive exploration through covariance matrix adaptation, and the order and/or stages in which degrees-of-freedom are freed is not pre-defined, as in for instance~\cite{Berthouze04,Bongard10,baranes11interaction}.


\section{Analysis: Individual Joints}

In this section, we analyze how and why PDFF arises when applying stochastic optimization, and how and why this depends on the arm structure.
We perform the analysis in a static context -- static because we do not perturb the parameters of the policy that determines the joint angles over time, but rather perturb the joint angles directly without a temporal component.
This analysis helps to understand why PDFF arises within an optimization context.


Here, we consider the effect that perturbations of individual joints have on the cost through sensitivity analysis.
Sensitivity analysis aims at \emph{``providing an understanding of how the model response variables respond to changes in the input.''}~\cite{saltelli00sensitivity}. We use sensitivity analysis to investigate how the variation in individual joint angles -- the input -- influences the variation in the cost -- the response variables. This provides a first indication of why PDFF arises. 

\subsection{Methods}
\begin{figure}[ht]
  \centering
  \includegraphics[width=0.6\columnwidth]{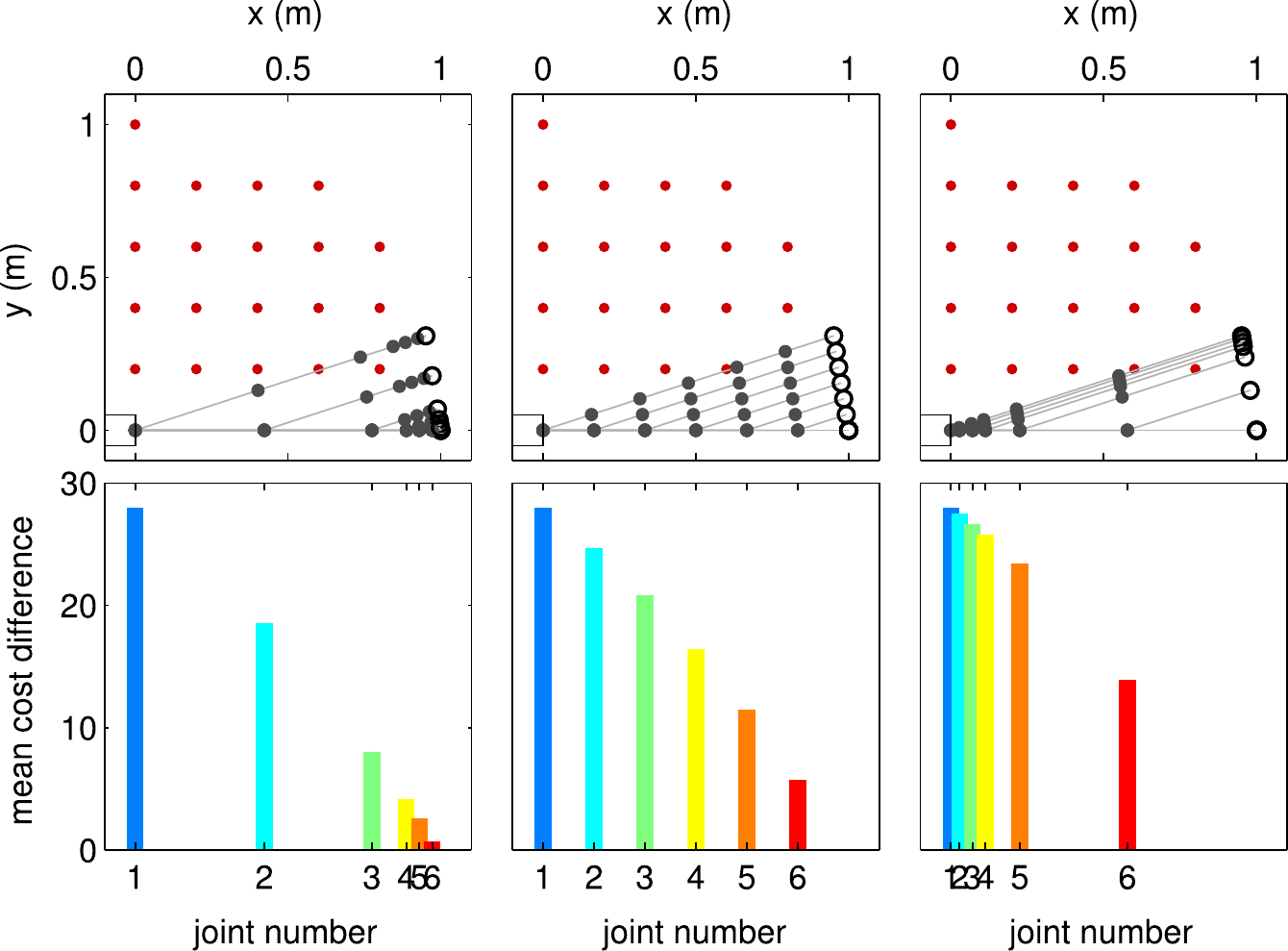}
  \caption{\label{fig_sensitivity_analysis} Results of the sensitivity analysis. For all arm configurations, we see that proximal joints lead to a higher average difference in the distance to the target than more distal ones.}
\end{figure}

In the default posture, all joint angles are zero. This posture is perturbed by setting one of the 6 joint angles to $\frac{\pi}{10}$. The 6 possible perturbations, one for each joint, are visualized in the top row of \figurename~\ref{fig_sensitivity_analysis}. For the default and perturbed configuration, we then compute the distance of the end-effector to the target $||\vx_{t_N}-\vx^g||$. Because this is a static context there are no joint accelerations, and the immediate costs (\ref{equ_cost_immediate}) are not included.
The lower row plots the difference in cost between the outstretched arm (where all joints are 0), and the slightly bent arm (where one joint angle is $\frac{\pi}{10}$). 
To acquire a value that is representative for the whole workspace, the differences in the lower row of \figurename~\ref{fig_sensitivity_analysis} is the average over the 20 target positions $\vx^g$ depicted in the top row of \figurename~\ref{fig_sensitivity_analysis}. 

\subsection{Results}

For all arm configurations, we see that proximal joints lead to a higher average difference in the distance to the target than more distal ones. This should not come as a surprise, as rotating more proximal joint leads to smaller movement in the end-effector space, and it is the end-effector space that determines the distance to the target. As a consequence, the same magnitude of perturbation will lead to a larger difference in cost for more proximal joints.

%

\subsection{Discussion}
The goal of stochastic optimization is to minimize costs through exploring and updating in parameter space. The results in \figurename~\ref{fig_sensitivity_analysis} demonstrate that perturbing proximal joints leads to larger differences in costs than distal joints. Therefore, an optimizer can be expected to minimize costs more quickly if it initially focuses exploration on proximal joints, rather than distal ones. 
This may be an explanation why exploration is larger in more proximal joints, but does not explain why distal joints are not also freed. This is the aim of our second analysis, which now follows.

\section{Analysis: Interactions Between Joints}

Whereas the previous section on sensitivity analysis considered joints \emph{individually}, we now turn to the \emph{interaction} between pairs of joints. We especially focus on how perturbations in proximal joints affect the influence of perturbations in more distal joints on the cost.

\subsection{Methods}
\begin{figure}[ht]
  \centering
  \includegraphics[width=0.8\columnwidth]{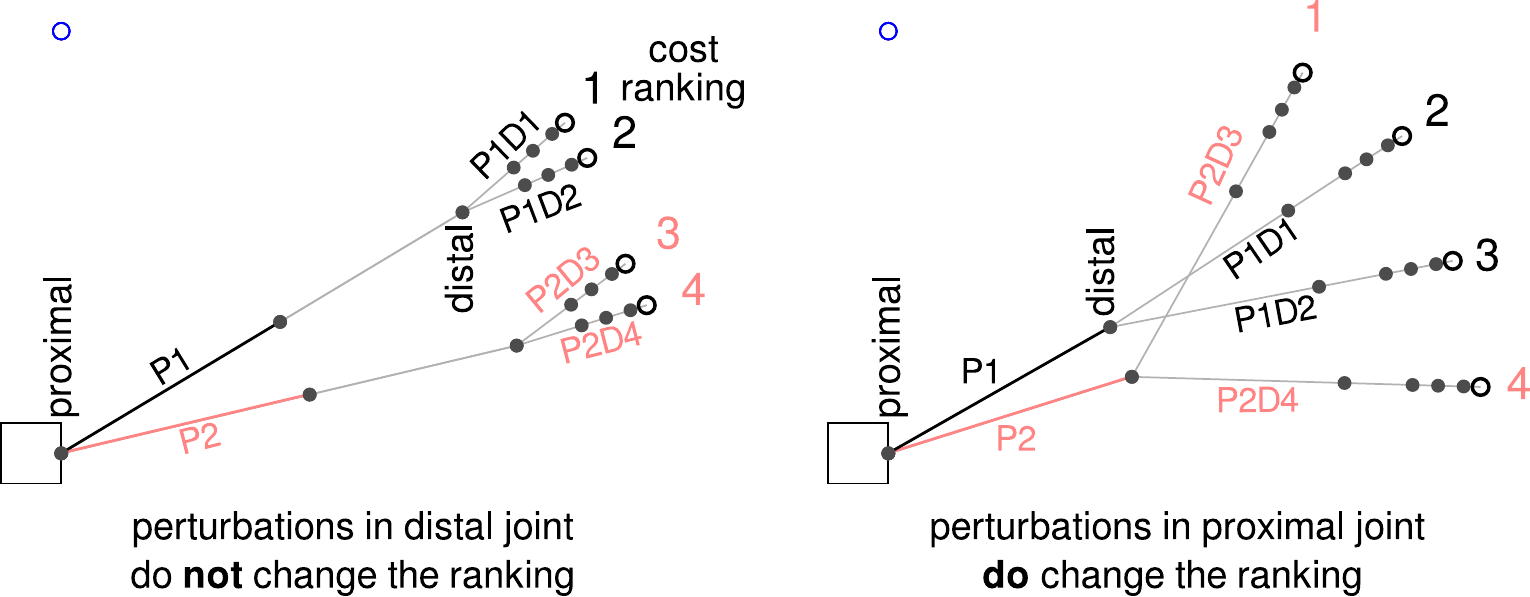}
  \caption{\label{fig_uncertainty_handling_example} Examples of perturbing pairs of joints for the human arm, and the effect of these perturbations on the cost ranking (numbers at the end of the arm).
  }
\end{figure}

Pairs of joints are considered. For the more proximal joint, two perturbations $\{P1,P2\}$ are sampled from $\N{0}{\frac{\pi}{10}}$. For each perturbation of the proximal joint, the distal joint is perturbed twice, also by sampling from  $\N{0}{\frac{\pi}{10}}$. This leads to the four arm configurations in \figurename~\ref{fig_uncertainty_handling_example}. The question we ask for these four configurations is: does the perturbation of the distal joint change the cost ranking? This is the same question underlying \emph{uncertainty handling} in ranked-based evolutionary direct policy search~\cite{heidrichmeisner08uncertainty}. \figurename~\ref{fig_uncertainty_handling_example} depicts examples where the answer is no (left) and yes (right). The question is asked for 100 different samples, and for each of the 20 target points. The average of these 2000 values represents the ratio that the answer was `no', i.e. a ratio of 1 implies that, no matter how much the distal joint is perturbed, it does not affect the ranking. A value of 0.5 implies that the perturbation of the distal joint affects the ranking half the time.

\subsection{Results}
\begin{figure}[ht]
  \centering
  \includegraphics[width=0.8\columnwidth]{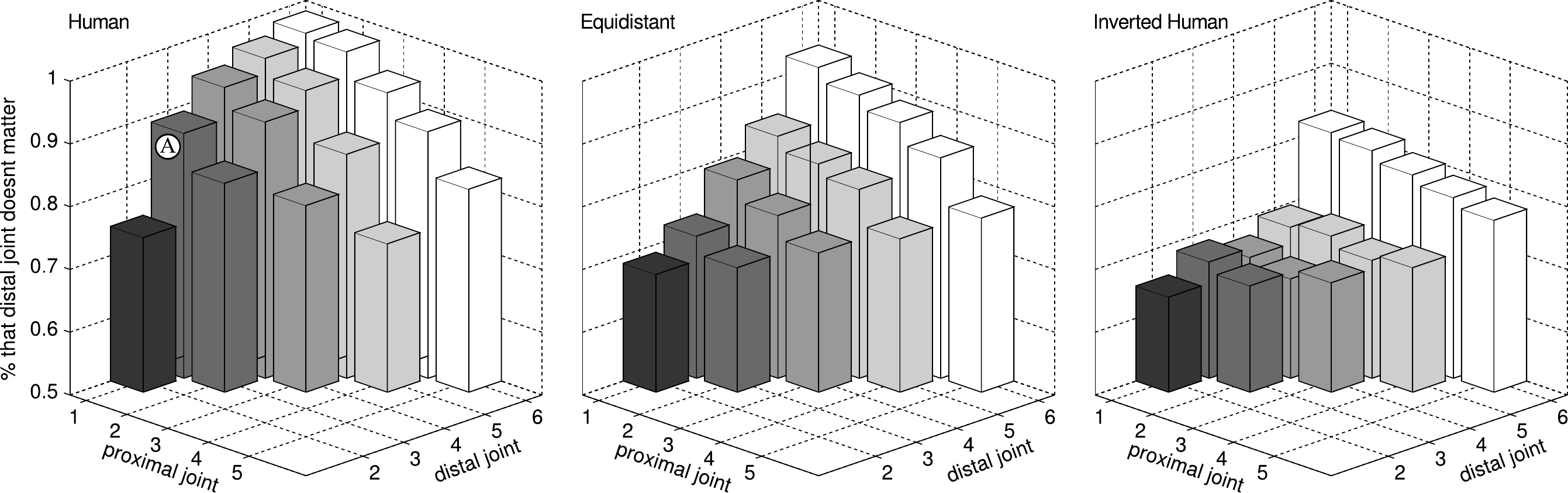}
  \caption{\label{fig_uncertainty_handling} For pairs of joints, the height of the bars represents the ratio that the perturbation of the distal joint does \emph{not} affect the cost ranking of the resulting postures.}
\end{figure}

\figurename~\ref{fig_uncertainty_handling} depicts the ratio for pairs of joints for all three arm configurations. For proximal joint 1 and distal joint 3 -- the case depicted in \figurename~\ref{fig_uncertainty_handling_example} -- this value is 0.89, labeled (A). Thus, when joint 1 is perturbed, the perturbation of joint 3 affect the cost ranking in only 11\% of the samples and target points. 

\subsection{Discussion}

The main implication of these results for stochastic optimization is that if we are performing exploration with for instance joint 1, exploration in more distal joints is less relevant to the cost. For the human arm for instance, it is not sensible to explore with joint 6 when exploring with joint 1, because joint 6 only affects the result 2\% of the time. Thus, from the point of view of stochastic optimization, joint 6 may well be frozen when searching in joint 1.

It is interesting to see that for the human arm configuration, proximal joints dominate distal joint much more (median=$0.89$) than an equidistant arm (median=$0.79$), and even more so for the `inverted'  human arm (median=$0.70$). Thus, exploring with more distal joints has a lower impact on the cost ranking for the human arm, and thus the effect of PDFF may be expected to be stronger for this configuration. This is confirmed by the empirical results in \figurename~\ref{fig_results}, where PDFF is most pronounced in the human arm configuration.


\section{General Discussion and Conclusion}

Staged motor learning, and in particular the progressive freezing and freeing of degrees of freedom observed in infant and adult motor exploration, has been argued to facilitate human acquisition of high-dimensional motor skills, and was also shown to be efficient for robot motor learning. Several hypotheses explaining the underlying mechanisms leading to such staged motor learning schedules were formulated so far, but have mostly relied on forms of innate scheduling of patterns of freeing and freezing. 

Here, in the framework of approximate optimal control, we have studied the hypothesis that staged learning schedules with freezing and progressive freeing of degrees of freedom can self-organize spontaneously as a result of the interaction between certain families of stochastic optimization methods with the physical properties of the body, and without involving physiological maturation. In particular, we have presented simulated experiments with a 6-DOF arm where a computational learner progressively acquired reaching skills, and we showed that a proximodistal organization appeared spontaneously. We also compared the emergent structuration as different arm structures are used -- from human-like to quite unnatural ones -- to study the effect of different kinematic structures on the emergence of PDFF. We analyzed further these results through a sensitivity analysis, providing a deeper understanding of \emph{why} PDFF arises during the stochastic optimization.

\textbf{Parsimony and biological plausibility} Overall, this model does not invalidate the hypothesis that an an innate maturational scheduling for freezing and freezing DOFs can be involved in infant motor learning. However, it shows that relatively simple stochastic optimization processes with adaptive exploration, which Bernstein already suggested were at play in infants, can already account for the formation of patterns of staged motor exploration. Yet, while the form of adaptive stochastic optimization we have considered is simple and general, one can wonder whether such mechanisms could be actually implemented in biologically plausible neural networks.
These mechanisms are a form of evolutionary optimization algorithms that are based on two complementary principles: the capacity to make variations/mutations of current good solutions (and to select the most useful ones), and the capacity to identify which directions of variation/mutation are currently most improving the current good solutions, involving a form of memory of past explorations. These two principles, and other more complex forms of Darwinian search processes in the brain, have been shown to be neurally plausible by several lines of research \cite{fernando2012selectionist}, building on Edelman theory of neuronal group selection \cite{edelman1987neural}, Changeux's theory of synaptic selection and selective stabilization \cite{changeux1973theory}, Calvin's replicating activity patterns \cite{calvin1987brain}. In particular, the recently developped Neuronal Replicator Hypothesis \cite{fernando2010neuronal} has shown how various known neuronal physiological mechanisms could implement such general genetic algorithms, including the mechanisms of adaptive exploration that we have been using in the model presented in this paper. An implementation of these mechanisms was shown to work with a neural network using realistic Izhikevich spiking neurons \cite{fernando2011evolvable}. To summarize, the mechanisms used to allow incremental exploration and learning in the model presented here are not only simple and parsimonious, but they are also implementable in realistic neural networks. 
 
\textbf{Open questions and research perspectives.}  
Our general hypothesis therefore forms a baseline against which more complex, domain-specific hypotheses should be compared. Also, the spontaneous formation of PDFF patterns through stochastic optimization appears to be compatible with observation of the patterns of motor exploration in adult motor learning, where maturational mechanisms have little probability to be at play. This new hypothesis also points to several open questions and new experimental investigations. In adults, several experiments have shown that a structuration of exploration through freezing and freezing of degrees of freedom happened~\cite{southard1987changing,hodges2005changes,vereijken1992free}, however to our knowledge these experiments did not systematically study to what extent the freeing or freezing of particular degrees of freedom was correlated to their current usefulness in progressing towards a goal.
Also, they considered tasks such as skiing, soccer or racket skills that were culturally known by subjects and thus could involve other mechanisms such as imitation learning, complicating the analysis of the exploration strategies. In infants, it is also an open question to know whether the adaptive exploration mechanisms shown by adults are already at play, or whether they are not yet in place and exploration is rather controlled by maturational mechanisms such as myelination. 

To address these open questions, one could imagine using experimental setups in which human subjects have to learn and explore new sensorimotor mapping that are highly different from what they already know, and where the relation between degrees of freedom and their usefulness for progressing towards the goal can be controlled systematically. Miard et al.~\cite{miard2014new} have proposed an experimental setup that could be used with adults in this context (but was not used for these specific questions so far): subjects have to learn how to control an abstract visual shape on a screen by using movements of their body as measured by a 3D camera (and the mathematical form of the relation between body movements and the abstract visual shape can be changed systematically). Similar setups could be imagined for infants, taking inspiration from the famous Rovee-Collier task \cite{rovee1999development} (developped for studying other questions): infants' movements (arms, legs) could be tracked with sensors and used to control the intensity/frequency of a sound or color of a light, through a mapping where the relation between degrees of freedom and their impact on the sound/light could be systematically changed to cancel out possible effects of myelination in the structuration of exploration. For example, it could be possible to program an inverse proximodistal relationship between arm movements and the sound/light: movements of the tip of the arm could be made to have more impact on the sound/light than movements of the shoulder. In such a case, observing an infant exploration with a corresponding inverse proximodistal law would reinforce our hypothesis that adaptive exploration plays an important role, while observing a standard proximodistal exploration would invalidate the hypothesis that such a mechanism is present or can have a leading role in exploration. 

\textbf{Complementarity with other mechanisms} It is also an open question to undertand how such adaptive exploration based on stochastic optimization could interact with other mechanisms guiding exploration such as maturation through myelination, imitation/social guidance, and intrinsic motivation. Several models in robotics have began to explore these links. Baranes and Oudeyer \cite{baranes11interaction} have studied the efficiency of combining stochastic optimization to reach goals with maturational mechanism which progressively grow the limits within which stochastic optimization can physically explore, showing an increase in efficiency from a machine learning point of view. Several works have shown how human demonstration of movements could bootstrap this optimization process (e.g. \cite{stulp14simultaneous}, or how humans can progressively shape subparts of the movements to complement autonomous exploration \cite{chernova2014robot}. Finally, exploration in infants is also highly driven by mechanisms of intrinsic motivation (also called curiosity), where instead of trying to reach a goal imposed by social peers or the experimenter (as in the model presented in this paper), they use intrinsic criteria such as information gain or surprise to set their own goals and choose how to practice these self-selected goals \cite{gottlieb2013information, 10.3389/fpsyg.2013.01006}. An important side effect of exploring multiple self-selected goals is that transfer learning across goals happens and in turn can shape the selection of future goals and ways to use degrees of freedom to explore them. Several computational architectures of intrinsically motivated learning have used stochastic optimization as the lower-level mechanism to learn how to reach self-selected goals in sensorimotor learning and achieve transfer learning across goals \cite{baranes:hal-00788440}. These integrated architecture have also shown the self-organization of developmental structure, such as the transition from non-articulated speech sounds to articulated vowels to proto-syllables in models if infant vocal development \cite{10.3389/fpsyg.2013.01006,TOPS:TOPS12196}.




\ifthenelse{\boolean{anonymous}}{
\input{13-Maturation-Stulp-anonymous.bbl}
}
{
\bibliographystyle{apacite}
\bibliography{bibliography-stulp,stulp-publications,12-ICDL-Stulp,additionalBib}

\begin{thebibliography}{}

\bibitem [\protect \citeauthoryear {%
Adolph%
\ \BBA {} Berger%
}{%
Adolph%
\ \BBA {} Berger%
}{%
{\protect \APACyear {2005}}%
}]{%
adolph2005physical}
\APACinsertmetastar {%
adolph2005physical}%
\begin{APACrefauthors}%
Adolph, K\BPBI E.%
\BCBT {}\ \BBA {} Berger, S\BPBI E.%
\end{APACrefauthors}%
\unskip\
\newblock
\APACrefYearMonthDay{2005}{}{}.
\newblock
{\BBOQ}\APACrefatitle {Physical and motor development} {Physical and motor
  development}.{\BBCQ}
\newblock
\APACjournalVolNumPages{Developmental science: An advanced
  textbook}{5}{}{223--281}.
\PrintBackRefs{\CurrentBib}

\bibitem [\protect \citeauthoryear {%
Arnold%
, Auger%
, Hansen%
\BCBL {}\ \BBA {} Ollivier%
}{%
Arnold%
\ \protect \BOthers {.}}{%
{\protect \APACyear {2011}}%
}]{%
arnold11informationgeometric}
\APACinsertmetastar {%
arnold11informationgeometric}%
\begin{APACrefauthors}%
Arnold, L.%
, Auger, A.%
, Hansen, N.%
\BCBL {}\ \BBA {} Ollivier, Y.%
\end{APACrefauthors}%
\unskip\
\newblock
\APACrefYearMonthDay{2011}{}{}.
\newblock
\APACrefbtitle {Information-Geometric Optimization Algorithms: A Unifying
  Picture via Invariance Principles} {Information-geometric optimization
  algorithms: A unifying picture via invariance principles}\
  \APACbVolEdTR{}{\BTR{}}.
\newblock
\APACaddressInstitution{}{INRIA Saclay}.
\PrintBackRefs{\CurrentBib}

\bibitem [\protect \citeauthoryear {%
Baranes%
\ \BBA {} Oudeyer%
}{%
Baranes%
\ \BBA {} Oudeyer%
}{%
{\protect \APACyear {2011}}%
}]{%
baranes11interaction}
\APACinsertmetastar {%
baranes11interaction}%
\begin{APACrefauthors}%
Baranes, A.%
\BCBT {}\ \BBA {} Oudeyer, P\BHBI Y.%
\end{APACrefauthors}%
\unskip\
\newblock
\APACrefYearMonthDay{2011}{}{}.
\newblock
{\BBOQ}\APACrefatitle {The Interaction of Maturational Constraints and
  Intrinsic Motivations in Active Motor Development} {The interaction of
  maturational constraints and intrinsic motivations in active motor
  development}.{\BBCQ}
\newblock
\BIn{} \APACrefbtitle {IEEE International Conference on Development and
  Learning.} {Ieee international conference on development and learning.}
\PrintBackRefs{\CurrentBib}

\bibitem [\protect \citeauthoryear {%
Baranes%
\ \BBA {} Oudeyer%
}{%
Baranes%
\ \BBA {} Oudeyer%
}{%
{\protect \APACyear {2013}}%
}]{%
baranes:hal-00788440}
\APACinsertmetastar {%
baranes:hal-00788440}%
\begin{APACrefauthors}%
Baranes, A.%
\BCBT {}\ \BBA {} Oudeyer, P\BHBI Y.%
\end{APACrefauthors}%
\unskip\
\newblock
\APACrefYearMonthDay{2013}{{\APACmonth{01}}}{}.
\newblock
{\BBOQ}\APACrefatitle {{Active Learning of Inverse Models with Intrinsically
  Motivated Goal Exploration in Robots}} {{Active Learning of Inverse Models
  with Intrinsically Motivated Goal Exploration in Robots}}.{\BBCQ}
\newblock
\APACjournalVolNumPages{{Robotics and Autonomous Systems}}{61}{1}{69-73}.
\newblock
\begin{APACrefURL} \url{http://hal.inria.fr/hal-00788440} \end{APACrefURL}
\newblock
\begin{APACrefDOI} \doi{10.1016/j.robot.2012.05.008} \end{APACrefDOI}
\PrintBackRefs{\CurrentBib}

\bibitem [\protect \citeauthoryear {%
Bernstein%
}{%
Bernstein%
}{%
{\protect \APACyear {1967}}%
}]{%
Bernstein67}
\APACinsertmetastar {%
Bernstein67}%
\begin{APACrefauthors}%
Bernstein, N.%
\end{APACrefauthors}%
\unskip\
\newblock
\APACrefYear{1967}.
\newblock
\APACrefbtitle {The Coordination and Regulation of Movements} {The coordination
  and regulation of movements}.
\newblock
\APACaddressPublisher{}{Pergamon}.
\PrintBackRefs{\CurrentBib}

\bibitem [\protect \citeauthoryear {%
Bertenthal%
\ \BBA {} von Hofsten%
}{%
Bertenthal%
\ \BBA {} von Hofsten%
}{%
{\protect \APACyear {1998}}%
}]{%
Bertenthal98}
\APACinsertmetastar {%
Bertenthal98}%
\begin{APACrefauthors}%
Bertenthal, B\BPBI I.%
\BCBT {}\ \BBA {} von Hofsten, C.%
\end{APACrefauthors}%
\unskip\
\newblock
\APACrefYearMonthDay{1998}{}{}.
\newblock
{\BBOQ}\APACrefatitle {Eye, head and trunk control: The foundation for manual
  development} {Eye, head and trunk control: The foundation for manual
  development}.{\BBCQ}
\newblock
\APACjournalVolNumPages{Neuroscience and Biobehavioral Review}{22}{}{515-520}.
\PrintBackRefs{\CurrentBib}

\bibitem [\protect \citeauthoryear {%
Berthier%
, Clifton%
, McCall%
\BCBL {}\ \BBA {} Robin%
}{%
Berthier%
\ \protect \BOthers {.}}{%
{\protect \APACyear {1999}}%
}]{%
Berthier99}
\APACinsertmetastar {%
Berthier99}%
\begin{APACrefauthors}%
Berthier, N\BPBI E.%
, Clifton, R.%
, McCall, D.%
\BCBL {}\ \BBA {} Robin, D.%
\end{APACrefauthors}%
\unskip\
\newblock
\APACrefYearMonthDay{1999}{}{}.
\newblock
{\BBOQ}\APACrefatitle {Proximodistal structure of early reaching in human
  infants} {Proximodistal structure of early reaching in human infants}.{\BBCQ}
\newblock
\APACjournalVolNumPages{Exp Brain Res}{}{}{}.
\PrintBackRefs{\CurrentBib}

\bibitem [\protect \citeauthoryear {%
Berthier%
, Rosenstein%
\BCBL {}\ \BBA {} Barto%
}{%
Berthier%
\ \protect \BOthers {.}}{%
{\protect \APACyear {2005}}%
}]{%
berthier2005approximate}
\APACinsertmetastar {%
berthier2005approximate}%
\begin{APACrefauthors}%
Berthier, N\BPBI E.%
, Rosenstein, M\BPBI T.%
\BCBL {}\ \BBA {} Barto, A\BPBI G.%
\end{APACrefauthors}%
\unskip\
\newblock
\APACrefYearMonthDay{2005}{}{}.
\newblock
{\BBOQ}\APACrefatitle {Approximate optimal control as a model for motor
  learning.} {Approximate optimal control as a model for motor
  learning.}{\BBCQ}
\newblock
\APACjournalVolNumPages{Psychological review}{112}{2}{329}.
\PrintBackRefs{\CurrentBib}

\bibitem [\protect \citeauthoryear {%
Berthouze%
\ \BBA {} Lungarella%
}{%
Berthouze%
\ \BBA {} Lungarella%
}{%
{\protect \APACyear {2004}}%
}]{%
Berthouze04}
\APACinsertmetastar {%
Berthouze04}%
\begin{APACrefauthors}%
Berthouze, L.%
\BCBT {}\ \BBA {} Lungarella, M.%
\end{APACrefauthors}%
\unskip\
\newblock
\APACrefYearMonthDay{2004}{}{}.
\newblock
{\BBOQ}\APACrefatitle {Motor Skill Acquisition under Environmental
  Perturbations: On the Necessity of Alternate Freezing and Freeing Degrees of
  Freedom} {Motor skill acquisition under environmental perturbations: On the
  necessity of alternate freezing and freeing degrees of freedom}.{\BBCQ}
\newblock
\APACjournalVolNumPages{Adaptive Behavior}{12}{1}{47-63}.
\PrintBackRefs{\CurrentBib}

\bibitem [\protect \citeauthoryear {%
Bongard%
}{%
Bongard%
}{%
{\protect \APACyear {2010}}%
}]{%
Bongard10}
\APACinsertmetastar {%
Bongard10}%
\begin{APACrefauthors}%
Bongard, J\BPBI C.%
\end{APACrefauthors}%
\unskip\
\newblock
\APACrefYearMonthDay{2010}{January}{}.
\newblock
{\BBOQ}\APACrefatitle {Morphological Change in Machines Accelerates the
  Evolution of Robust Behavior} {Morphological change in machines accelerates
  the evolution of robust behavior}.{\BBCQ}
\newblock
\APACjournalVolNumPages{Proceedigns of the National Academy of Sciences of the
  United States of America (PNAS)}{}{}{}.
\PrintBackRefs{\CurrentBib}

\bibitem [\protect \citeauthoryear {%
Calvin%
}{%
Calvin%
}{%
{\protect \APACyear {1987}}%
}]{%
calvin1987brain}
\APACinsertmetastar {%
calvin1987brain}%
\begin{APACrefauthors}%
Calvin, W\BPBI H.%
\end{APACrefauthors}%
\unskip\
\newblock
\APACrefYearMonthDay{1987}{}{}.
\newblock
{\BBOQ}\APACrefatitle {The brain as a Darwin machine} {The brain as a darwin
  machine}.{\BBCQ}
\newblock
\APACjournalVolNumPages{Nature}{330}{}{33--34}.
\PrintBackRefs{\CurrentBib}

\bibitem [\protect \citeauthoryear {%
Cangelosi%
}{%
Cangelosi%
}{%
{\protect \APACyear {1999}}%
}]{%
Cangelosi99}
\APACinsertmetastar {%
Cangelosi99}%
\begin{APACrefauthors}%
Cangelosi, A.%
\end{APACrefauthors}%
\unskip\
\newblock
\APACrefYearMonthDay{1999}{}{}.
\newblock
{\BBOQ}\APACrefatitle {Heterochrony and adaptation in developing neural
  networks} {Heterochrony and adaptation in developing neural networks}.{\BBCQ}
\newblock
\BIn{} W\BPBI B.~et al.\ (\BED), \APACrefbtitle {Proceedings of the Genetic and
  Evolutionary Computation Conference} {Proceedings of the genetic and
  evolutionary computation conference}\ (\BPGS\ 1241--1248).
\newblock
\APACaddressPublisher{}{San Francisco, CA: Morgan Kaufmann}.
\PrintBackRefs{\CurrentBib}

\bibitem [\protect \citeauthoryear {%
Changeux%
, Courr{\'e}ge%
\BCBL {}\ \BBA {} Danchin%
}{%
Changeux%
\ \protect \BOthers {.}}{%
{\protect \APACyear {1973}}%
}]{%
changeux1973theory}
\APACinsertmetastar {%
changeux1973theory}%
\begin{APACrefauthors}%
Changeux, J\BHBI P.%
, Courr{\'e}ge, P.%
\BCBL {}\ \BBA {} Danchin, A.%
\end{APACrefauthors}%
\unskip\
\newblock
\APACrefYearMonthDay{1973}{}{}.
\newblock
{\BBOQ}\APACrefatitle {A theory of the epigenesis of neuronal networks by
  selective stabilization of synapses} {A theory of the epigenesis of neuronal
  networks by selective stabilization of synapses}.{\BBCQ}
\newblock
\APACjournalVolNumPages{Proceedings of the National Academy of
  Sciences}{70}{10}{2974--2978}.
\PrintBackRefs{\CurrentBib}

\bibitem [\protect \citeauthoryear {%
Chernova%
\ \BBA {} Thomaz%
}{%
Chernova%
\ \BBA {} Thomaz%
}{%
{\protect \APACyear {2014}}%
}]{%
chernova2014robot}
\APACinsertmetastar {%
chernova2014robot}%
\begin{APACrefauthors}%
Chernova, S.%
\BCBT {}\ \BBA {} Thomaz, A\BPBI L.%
\end{APACrefauthors}%
\unskip\
\newblock
\APACrefYearMonthDay{2014}{}{}.
\newblock
{\BBOQ}\APACrefatitle {Robot learning from human teachers} {Robot learning from
  human teachers}.{\BBCQ}
\newblock
\APACjournalVolNumPages{Synthesis Lectures on Artificial Intelligence and
  Machine Learning}{8}{3}{1--121}.
\PrintBackRefs{\CurrentBib}

\bibitem [\protect \citeauthoryear {%
Cohen%
\ \BBA {} Rosenbaum%
}{%
Cohen%
\ \BBA {} Rosenbaum%
}{%
{\protect \APACyear {2004}}%
}]{%
cohen04where}
\APACinsertmetastar {%
cohen04where}%
\begin{APACrefauthors}%
Cohen, R\BPBI G.%
\BCBT {}\ \BBA {} Rosenbaum, D\BPBI A.%
\end{APACrefauthors}%
\unskip\
\newblock
\APACrefYearMonthDay{2004}{}{}.
\newblock
{\BBOQ}\APACrefatitle {Where grasps are made reveals how grasps are planned:
  generation and recall of motor plans} {Where grasps are made reveals how
  grasps are planned: generation and recall of motor plans}.{\BBCQ}
\newblock
\APACjournalVolNumPages{Exp Brain Res}{157}{4}{486-495}.
\newblock
\begin{APACrefDOI} \doi{10.1007/s00221-004-1862-9} \end{APACrefDOI}
\PrintBackRefs{\CurrentBib}

\bibitem [\protect \citeauthoryear {%
Edelman%
}{%
Edelman%
}{%
{\protect \APACyear {1987}}%
}]{%
edelman1987neural}
\APACinsertmetastar {%
edelman1987neural}%
\begin{APACrefauthors}%
Edelman, G\BPBI M.%
\end{APACrefauthors}%
\unskip\
\newblock
\APACrefYear{1987}.
\newblock
\APACrefbtitle {Neural Darwinism: The theory of neuronal group selection.}
  {Neural darwinism: The theory of neuronal group selection.}
\newblock
\APACaddressPublisher{}{Basic Books}.
\PrintBackRefs{\CurrentBib}

\bibitem [\protect \citeauthoryear {%
Elman%
}{%
Elman%
}{%
{\protect \APACyear {1993}}%
}]{%
Elman93}
\APACinsertmetastar {%
Elman93}%
\begin{APACrefauthors}%
Elman, J.%
\end{APACrefauthors}%
\unskip\
\newblock
\APACrefYearMonthDay{1993}{}{}.
\newblock
{\BBOQ}\APACrefatitle {Learning and development in neural networks: The
  importance of starting small} {Learning and development in neural networks:
  The importance of starting small}.{\BBCQ}
\newblock
\APACjournalVolNumPages{Cognition}{48}{}{71-99}.
\PrintBackRefs{\CurrentBib}

\bibitem [\protect \citeauthoryear {%
C.~Fernando%
, Goldstein%
\BCBL {}\ \BBA {} Szathm{\'a}ry%
}{%
C.~Fernando%
\ \protect \BOthers {.}}{%
{\protect \APACyear {2010}}%
}]{%
fernando2010neuronal}
\APACinsertmetastar {%
fernando2010neuronal}%
\begin{APACrefauthors}%
Fernando, C.%
, Goldstein, R.%
\BCBL {}\ \BBA {} Szathm{\'a}ry, E.%
\end{APACrefauthors}%
\unskip\
\newblock
\APACrefYearMonthDay{2010}{}{}.
\newblock
{\BBOQ}\APACrefatitle {The neuronal replicator hypothesis} {The neuronal
  replicator hypothesis}.{\BBCQ}
\newblock
\APACjournalVolNumPages{Neural computation}{22}{11}{2809--2857}.
\PrintBackRefs{\CurrentBib}

\bibitem [\protect \citeauthoryear {%
C.~Fernando%
, Vasas%
, Szathm{\'a}ry%
\BCBL {}\ \BBA {} Husbands%
}{%
C.~Fernando%
\ \protect \BOthers {.}}{%
{\protect \APACyear {2011}}%
}]{%
fernando2011evolvable}
\APACinsertmetastar {%
fernando2011evolvable}%
\begin{APACrefauthors}%
Fernando, C.%
, Vasas, V.%
, Szathm{\'a}ry, E.%
\BCBL {}\ \BBA {} Husbands, P.%
\end{APACrefauthors}%
\unskip\
\newblock
\APACrefYearMonthDay{2011}{}{}.
\newblock
{\BBOQ}\APACrefatitle {Evolvable neuronal paths: a novel basis for information
  and search in the brain} {Evolvable neuronal paths: a novel basis for
  information and search in the brain}.{\BBCQ}
\newblock
\APACjournalVolNumPages{PloS one}{6}{8}{e23534}.
\PrintBackRefs{\CurrentBib}

\bibitem [\protect \citeauthoryear {%
C\BPBI T.~Fernando%
, Szathmary%
\BCBL {}\ \BBA {} Husbands%
}{%
C\BPBI T.~Fernando%
\ \protect \BOthers {.}}{%
{\protect \APACyear {2012}}%
}]{%
fernando2012selectionist}
\APACinsertmetastar {%
fernando2012selectionist}%
\begin{APACrefauthors}%
Fernando, C\BPBI T.%
, Szathmary, E.%
\BCBL {}\ \BBA {} Husbands, P.%
\end{APACrefauthors}%
\unskip\
\newblock
\APACrefYearMonthDay{2012}{}{}.
\newblock
{\BBOQ}\APACrefatitle {Selectionist and evolutionary approaches to brain
  function: a critical appraisal} {Selectionist and evolutionary approaches to
  brain function: a critical appraisal}.{\BBCQ}
\newblock
\APACjournalVolNumPages{Frontiers in computational neuroscience}{6}{}{24}.
\PrintBackRefs{\CurrentBib}

\bibitem [\protect \citeauthoryear {%
French%
, Mermillod%
, Quinn%
, Chauvin%
\BCBL {}\ \BBA {} Mareschal%
}{%
French%
\ \protect \BOthers {.}}{%
{\protect \APACyear {2002}}%
}]{%
French02}
\APACinsertmetastar {%
French02}%
\begin{APACrefauthors}%
French, R\BPBI M.%
, Mermillod, M.%
, Quinn, P\BPBI C.%
, Chauvin, A.%
\BCBL {}\ \BBA {} Mareschal, D.%
\end{APACrefauthors}%
\unskip\
\newblock
\APACrefYearMonthDay{2002}{}{}.
\newblock
{\BBOQ}\APACrefatitle {The importance of starting blurry: Simulating improved
  basic-level category learning in infants due to weak visual acuity} {The
  importance of starting blurry: Simulating improved basic-level category
  learning in infants due to weak visual acuity}.{\BBCQ}
\newblock
\BIn{} LEA\ (\BED), \APACrefbtitle {Proceedings of the 24th Annual Conference
  of the Cognitive Science Society} {Proceedings of the 24th annual conference
  of the cognitive science society}\ (\BPG~322-327).
\newblock
\APACaddressPublisher{New Jersey}{}.
\PrintBackRefs{\CurrentBib}

\bibitem [\protect \citeauthoryear {%
Gottlieb%
, Oudeyer%
, Lopes%
\BCBL {}\ \BBA {} Baranes%
}{%
Gottlieb%
\ \protect \BOthers {.}}{%
{\protect \APACyear {2013}}%
}]{%
gottlieb2013information}
\APACinsertmetastar {%
gottlieb2013information}%
\begin{APACrefauthors}%
Gottlieb, J.%
, Oudeyer, P\BHBI Y.%
, Lopes, M.%
\BCBL {}\ \BBA {} Baranes, A.%
\end{APACrefauthors}%
\unskip\
\newblock
\APACrefYearMonthDay{2013}{}{}.
\newblock
{\BBOQ}\APACrefatitle {Information-seeking, curiosity, and attention:
  computational and neural mechanisms} {Information-seeking, curiosity, and
  attention: computational and neural mechanisms}.{\BBCQ}
\newblock
\APACjournalVolNumPages{Trends in cognitive sciences}{17}{11}{585--593}.
\PrintBackRefs{\CurrentBib}

\bibitem [\protect \citeauthoryear {%
Grupen%
}{%
Grupen%
}{%
{\protect \APACyear {2003}}%
}]{%
Grupen03}
\APACinsertmetastar {%
Grupen03}%
\begin{APACrefauthors}%
Grupen, R.%
\end{APACrefauthors}%
\unskip\
\newblock
\APACrefYearMonthDay{2003}{August}{}.
\newblock
{\BBOQ}\APACrefatitle {A Developmental Organization for Robot Behavior} {A
  developmental organization for robot behavior}.{\BBCQ}
\newblock
\BIn{} \APACrefbtitle {Proceedings of the third International Workshop on
  Epigenenetic Robotics.} {Proceedings of the third international workshop on
  epigenenetic robotics.}
\newblock
\APACaddressPublisher{Boston, MA}{}.
\PrintBackRefs{\CurrentBib}

\bibitem [\protect \citeauthoryear {%
Hansen%
\ \BBA {} Ostermeier%
}{%
Hansen%
\ \BBA {} Ostermeier%
}{%
{\protect \APACyear {2001}}%
}]{%
hansen01completely}
\APACinsertmetastar {%
hansen01completely}%
\begin{APACrefauthors}%
Hansen, N.%
\BCBT {}\ \BBA {} Ostermeier, A.%
\end{APACrefauthors}%
\unskip\
\newblock
\APACrefYearMonthDay{2001}{}{}.
\newblock
{\BBOQ}\APACrefatitle {Completely derandomized self-adaptation in evolution
  strategies} {Completely derandomized self-adaptation in evolution
  strategies}.{\BBCQ}
\newblock
\APACjournalVolNumPages{Evolutionary Computation}{9}{2}{159-195}.
\PrintBackRefs{\CurrentBib}

\bibitem [\protect \citeauthoryear {%
Heidrich-Meisner%
\ \BBA {} Igel%
}{%
Heidrich-Meisner%
\ \BBA {} Igel%
}{%
{\protect \APACyear {2008}}%
}]{%
heidrichmeisner08uncertainty}
\APACinsertmetastar {%
heidrichmeisner08uncertainty}%
\begin{APACrefauthors}%
Heidrich-Meisner, V.%
\BCBT {}\ \BBA {} Igel, C.%
\end{APACrefauthors}%
\unskip\
\newblock
\APACrefYearMonthDay{2008}{}{}.
\newblock
{\BBOQ}\APACrefatitle {Uncertainty Handling in Evolutionary Direct Policy
  Search} {Uncertainty handling in evolutionary direct policy search}.{\BBCQ}.
\PrintBackRefs{\CurrentBib}

\bibitem [\protect \citeauthoryear {%
Hodges%
, Hayes%
, Horn%
\BCBL {}\ \BBA {} Williams%
}{%
Hodges%
\ \protect \BOthers {.}}{%
{\protect \APACyear {2005}}%
}]{%
hodges2005changes}
\APACinsertmetastar {%
hodges2005changes}%
\begin{APACrefauthors}%
Hodges, N\BPBI J.%
, Hayes, S.%
, Horn, R\BPBI R.%
\BCBL {}\ \BBA {} Williams, A\BPBI M.%
\end{APACrefauthors}%
\unskip\
\newblock
\APACrefYearMonthDay{2005}{}{}.
\newblock
{\BBOQ}\APACrefatitle {Changes in coordination, control and outcome as a result
  of extended practice on a novel motor skill} {Changes in coordination,
  control and outcome as a result of extended practice on a novel motor
  skill}.{\BBCQ}
\newblock
\APACjournalVolNumPages{Ergonomics}{48}{11-14}{1672--1685}.
\PrintBackRefs{\CurrentBib}

\bibitem [\protect \citeauthoryear {%
Hoffmann%
, Theodorou%
\BCBL {}\ \BBA {} Schaal%
}{%
Hoffmann%
\ \protect \BOthers {.}}{%
{\protect \APACyear {2008}}%
}]{%
hoffmann08optimization}
\APACinsertmetastar {%
hoffmann08optimization}%
\begin{APACrefauthors}%
Hoffmann, H.%
, Theodorou, E.%
\BCBL {}\ \BBA {} Schaal, S.%
\end{APACrefauthors}%
\unskip\
\newblock
\APACrefYearMonthDay{2008}{}{}.
\newblock
{\BBOQ}\APACrefatitle {Optimization strategies in human reinforcement learning}
  {Optimization strategies in human reinforcement learning}.{\BBCQ}
\newblock
\APACjournalVolNumPages{Advances in computational motor control VII.
  Washington, DC: Society for Neuroscience}{}{}{}.
\PrintBackRefs{\CurrentBib}

\bibitem [\protect \citeauthoryear {%
Jansen%
\ \BBA {} Fladby%
}{%
Jansen%
\ \BBA {} Fladby%
}{%
{\protect \APACyear {1990}}%
}]{%
jansen1990perinatal}
\APACinsertmetastar {%
jansen1990perinatal}%
\begin{APACrefauthors}%
Jansen, J.%
\BCBT {}\ \BBA {} Fladby, T.%
\end{APACrefauthors}%
\unskip\
\newblock
\APACrefYearMonthDay{1990}{}{}.
\newblock
{\BBOQ}\APACrefatitle {The perinatal reorganization of the innervation of
  skeletal muscle in mammals} {The perinatal reorganization of the innervation
  of skeletal muscle in mammals}.{\BBCQ}
\newblock
\APACjournalVolNumPages{Progress in neurobiology}{34}{1}{39--90}.
\PrintBackRefs{\CurrentBib}

\bibitem [\protect \citeauthoryear {%
Kober%
\ \BBA {} Peters%
}{%
Kober%
\ \BBA {} Peters%
}{%
{\protect \APACyear {2011}}%
}]{%
kober11policy}
\APACinsertmetastar {%
kober11policy}%
\begin{APACrefauthors}%
Kober, J.%
\BCBT {}\ \BBA {} Peters, J.%
\end{APACrefauthors}%
\unskip\
\newblock
\APACrefYearMonthDay{2011}{}{}.
\newblock
{\BBOQ}\APACrefatitle {Policy Search for Motor Primitives in Robotics} {Policy
  search for motor primitives in robotics}.{\BBCQ}
\newblock
\APACjournalVolNumPages{Machine Learning}{84}{}{171-203}.
\PrintBackRefs{\CurrentBib}

\bibitem [\protect \citeauthoryear {%
Kuypers%
}{%
Kuypers%
}{%
{\protect \APACyear {1981}}%
}]{%
kuypers1981anatomy}
\APACinsertmetastar {%
kuypers1981anatomy}%
\begin{APACrefauthors}%
Kuypers, H.%
\end{APACrefauthors}%
\unskip\
\newblock
\APACrefYearMonthDay{1981}{}{}.
\newblock
{\BBOQ}\APACrefatitle {Anatomy of the descending pathways} {Anatomy of the
  descending pathways}.{\BBCQ}
\newblock
\APACjournalVolNumPages{Comprehensive Physiology}{}{}{}.
\PrintBackRefs{\CurrentBib}

\bibitem [\protect \citeauthoryear {%
Lee%
, Meng%
\BCBL {}\ \BBA {} Chao%
}{%
Lee%
\ \protect \BOthers {.}}{%
{\protect \APACyear {2007}}%
}]{%
Lee07b}
\APACinsertmetastar {%
Lee07b}%
\begin{APACrefauthors}%
Lee, M.%
, Meng, Q.%
\BCBL {}\ \BBA {} Chao, F.%
\end{APACrefauthors}%
\unskip\
\newblock
\APACrefYearMonthDay{2007}{}{}.
\newblock
{\BBOQ}\APACrefatitle {Developmental Learning for Autonomous Robots}
  {Developmental learning for autonomous robots}.{\BBCQ}
\newblock
\APACjournalVolNumPages{Robotics and Autonomous Systems}{55}{9}{750-759}.
\PrintBackRefs{\CurrentBib}

\bibitem [\protect \citeauthoryear {%
Matos%
, Suzuki%
\BCBL {}\ \BBA {} Arita%
}{%
Matos%
\ \protect \BOthers {.}}{%
{\protect \APACyear {2007}}%
}]{%
Matos07}
\APACinsertmetastar {%
Matos07}%
\begin{APACrefauthors}%
Matos, A.%
, Suzuki, R.%
\BCBL {}\ \BBA {} Arita, T.%
\end{APACrefauthors}%
\unskip\
\newblock
\APACrefYearMonthDay{2007}{}{}.
\newblock
{\BBOQ}\APACrefatitle {Heterochrony and evolvability in neural network
  development} {Heterochrony and evolvability in neural network
  development}.{\BBCQ}
\newblock
\APACjournalVolNumPages{Artificial Life Robotics}{11}{}{175-182}.
\PrintBackRefs{\CurrentBib}

\bibitem [\protect \citeauthoryear {%
Miard%
\ \protect \BOthers {.}}{%
Miard%
\ \protect \BOthers {.}}{%
{\protect \APACyear {2014}}%
}]{%
miard2014new}
\APACinsertmetastar {%
miard2014new}%
\begin{APACrefauthors}%
Miard, B.%
, Rouanet, P.%
, Grizou, J.%
, Lopes, M.%
, Gottlieb, J.%
, Baranes, A.%
\BCBL {}\ \BBA {} Oudeyer, P\BHBI Y.%
\end{APACrefauthors}%
\unskip\
\newblock
\APACrefYearMonthDay{2014}{}{}.
\newblock
{\BBOQ}\APACrefatitle {A new experimental setup to study the structure of
  curiosity-driven exploration in humans} {A new experimental setup to study
  the structure of curiosity-driven exploration in humans}.{\BBCQ}
\newblock
\BIn{} \APACrefbtitle {Proceedings of ICDL-EPIROB 2014.} {Proceedings of
  icdl-epirob 2014.}
\PrintBackRefs{\CurrentBib}

\bibitem [\protect \citeauthoryear {%
Moulin-Frier%
, Nguyen%
\BCBL {}\ \BBA {} Oudeyer%
}{%
Moulin-Frier%
\ \protect \BOthers {.}}{%
{\protect \APACyear {2014}}%
}]{%
10.3389/fpsyg.2013.01006}
\APACinsertmetastar {%
10.3389/fpsyg.2013.01006}%
\begin{APACrefauthors}%
Moulin-Frier, C.%
, Nguyen, S\BPBI M.%
\BCBL {}\ \BBA {} Oudeyer, P\BHBI Y.%
\end{APACrefauthors}%
\unskip\
\newblock
\APACrefYearMonthDay{2014}{}{}.
\newblock
{\BBOQ}\APACrefatitle {Self-organization of early vocal development in infants
  and machines: the role of intrinsic motivation} {Self-organization of early
  vocal development in infants and machines: the role of intrinsic
  motivation}.{\BBCQ}
\newblock
\APACjournalVolNumPages{Frontiers in Psychology}{4}{}{1006}.
\newblock
\begin{APACrefURL}
  \url{http://journal.frontiersin.org/article/10.3389/fpsyg.2013.01006}
  \end{APACrefURL}
\newblock
\begin{APACrefDOI} \doi{10.3389/fpsyg.2013.01006} \end{APACrefDOI}
\PrintBackRefs{\CurrentBib}

\bibitem [\protect \citeauthoryear {%
Nagai%
, Asada%
\BCBL {}\ \BBA {} Hosoda%
}{%
Nagai%
\ \protect \BOthers {.}}{%
{\protect \APACyear {2006}}%
}]{%
Nagai06}
\APACinsertmetastar {%
Nagai06}%
\begin{APACrefauthors}%
Nagai, Y.%
, Asada, M.%
\BCBL {}\ \BBA {} Hosoda, K.%
\end{APACrefauthors}%
\unskip\
\newblock
\APACrefYearMonthDay{2006}{September}{}.
\newblock
{\BBOQ}\APACrefatitle {Learning for joint attention helped by functional
  development} {Learning for joint attention helped by functional
  development}.{\BBCQ}
\newblock
\APACjournalVolNumPages{Advanced Robotics}{20}{10}{1165-1181}.
\PrintBackRefs{\CurrentBib}

\bibitem [\protect \citeauthoryear {%
Oudeyer%
, Baranes%
\BCBL {}\ \BBA {} Kaplan%
}{%
Oudeyer%
\ \protect \BOthers {.}}{%
{\protect \APACyear {2013}}%
}]{%
oudeyer:hal-00788611}
\APACinsertmetastar {%
oudeyer:hal-00788611}%
\begin{APACrefauthors}%
Oudeyer, P\BHBI Y.%
, Baranes, A.%
\BCBL {}\ \BBA {} Kaplan, F.%
\end{APACrefauthors}%
\unskip\
\newblock
\APACrefYearMonthDay{2013}{{\APACmonth{02}}}{}.
\newblock
{\BBOQ}\APACrefatitle {{Intrinsically Motivated Learning of Real World
  Sensorimotor Skills with Developmental Constraints}} {{Intrinsically
  Motivated Learning of Real World Sensorimotor Skills with Developmental
  Constraints}}.{\BBCQ}
\newblock
\BIn{} G.~Baldassarre\ \BBA {} M.~Mirolli\ (\BEDS), \APACrefbtitle
  {{Intrinsically Motivated Learning in Natural and Artificial Systems}.}
  {{Intrinsically Motivated Learning in Natural and Artificial Systems}.}
\newblock
\APACaddressPublisher{}{Springer}.
\newblock
\begin{APACrefURL} \url{http://hal.inria.fr/hal-00788611} \end{APACrefURL}
\PrintBackRefs{\CurrentBib}

\bibitem [\protect \citeauthoryear {%
Oudeyer%
\ \BBA {} Smith%
}{%
Oudeyer%
\ \BBA {} Smith%
}{%
{\protect \APACyear {2016}}%
}]{%
TOPS:TOPS12196}
\APACinsertmetastar {%
TOPS:TOPS12196}%
\begin{APACrefauthors}%
Oudeyer, P\BHBI Y.%
\BCBT {}\ \BBA {} Smith, L\BPBI B.%
\end{APACrefauthors}%
\unskip\
\newblock
\APACrefYearMonthDay{2016}{}{}.
\newblock
{\BBOQ}\APACrefatitle {How Evolution May Work Through Curiosity-Driven
  Developmental Process} {How evolution may work through curiosity-driven
  developmental process}.{\BBCQ}
\newblock
\APACjournalVolNumPages{Topics in Cognitive Science}{8}{2}{492--502}.
\newblock
\begin{APACrefURL} \url{http://dx.doi.org/10.1111/tops.12196} \end{APACrefURL}
\newblock
\begin{APACrefDOI} \doi{10.1111/tops.12196} \end{APACrefDOI}
\PrintBackRefs{\CurrentBib}

\bibitem [\protect \citeauthoryear {%
Rovee-Collier%
}{%
Rovee-Collier%
}{%
{\protect \APACyear {1999}}%
}]{%
rovee1999development}
\APACinsertmetastar {%
rovee1999development}%
\begin{APACrefauthors}%
Rovee-Collier, C.%
\end{APACrefauthors}%
\unskip\
\newblock
\APACrefYearMonthDay{1999}{}{}.
\newblock
{\BBOQ}\APACrefatitle {The development of infant memory} {The development of
  infant memory}.{\BBCQ}
\newblock
\APACjournalVolNumPages{Current Directions in Psychological
  Science}{8}{3}{80--85}.
\PrintBackRefs{\CurrentBib}

\bibitem [\protect \citeauthoryear {%
Rubinstein%
\ \BBA {} Kroese%
}{%
Rubinstein%
\ \BBA {} Kroese%
}{%
{\protect \APACyear {2004}}%
}]{%
rubinstein04crossentropy}
\APACinsertmetastar {%
rubinstein04crossentropy}%
\begin{APACrefauthors}%
Rubinstein, R.%
\BCBT {}\ \BBA {} Kroese, D.%
\end{APACrefauthors}%
\unskip\
\newblock
\APACrefYear{2004}.
\newblock
\APACrefbtitle {The Cross-Entropy Method: A Unified Approach to Combinatorial
  Optimization, Monte-Carlo Simulation, and Machine Learning} {The
  cross-entropy method: A unified approach to combinatorial optimization,
  monte-carlo simulation, and machine learning}.
\newblock
\APACaddressPublisher{}{Springer-Verlag}.
\PrintBackRefs{\CurrentBib}

\bibitem [\protect \citeauthoryear {%
Sakoe%
\ \BBA {} Chiba%
}{%
Sakoe%
\ \BBA {} Chiba%
}{%
{\protect \APACyear {1978}}%
}]{%
sakoe1978dynamic}
\APACinsertmetastar {%
sakoe1978dynamic}%
\begin{APACrefauthors}%
Sakoe, H.%
\BCBT {}\ \BBA {} Chiba, S.%
\end{APACrefauthors}%
\unskip\
\newblock
\APACrefYearMonthDay{1978}{}{}.
\newblock
{\BBOQ}\APACrefatitle {{Dynamic programming algorithm optimization for spoken
  word recognition}} {{Dynamic programming algorithm optimization for spoken
  word recognition}}.{\BBCQ}
\newblock
\APACjournalVolNumPages{IEEE Trans. on Acoustics, Speech, and Signal
  Processing}{26}{}{43--49}.
\PrintBackRefs{\CurrentBib}

\bibitem [\protect \citeauthoryear {%
Saltelli%
, Chan%
\BCBL {}\ \BBA {} Scott%
}{%
Saltelli%
\ \protect \BOthers {.}}{%
{\protect \APACyear {2000}}%
}]{%
saltelli00sensitivity}
\APACinsertmetastar {%
saltelli00sensitivity}%
\begin{APACrefauthors}%
Saltelli, A.%
, Chan, K.%
\BCBL {}\ \BBA {} Scott, E\BPBI M.%
\end{APACrefauthors}%
\unskip\
\newblock
\APACrefYear{2000}.
\newblock
\APACrefbtitle {{S}ensitivity analysis} {{S}ensitivity analysis}.
\newblock
\APACaddressPublisher{}{Chichester: Wiley}.
\PrintBackRefs{\CurrentBib}

\bibitem [\protect \citeauthoryear {%
Schlesinger%
, Parisi%
\BCBL {}\ \BBA {} Langer%
}{%
Schlesinger%
\ \protect \BOthers {.}}{%
{\protect \APACyear {2000}}%
}]{%
Schlesinger00}
\APACinsertmetastar {%
Schlesinger00}%
\begin{APACrefauthors}%
Schlesinger, M.%
, Parisi, D.%
\BCBL {}\ \BBA {} Langer, J.%
\end{APACrefauthors}%
\unskip\
\newblock
\APACrefYearMonthDay{2000}{}{}.
\newblock
{\BBOQ}\APACrefatitle {Learning to reach by constraining the movement search
  space} {Learning to reach by constraining the movement search space}.{\BBCQ}
\newblock
\APACjournalVolNumPages{Developmental Science}{3}{}{67-80}.
\PrintBackRefs{\CurrentBib}

\bibitem [\protect \citeauthoryear {%
Southard%
\ \BBA {} Higgins%
}{%
Southard%
\ \BBA {} Higgins%
}{%
{\protect \APACyear {1987}}%
}]{%
southard1987changing}
\APACinsertmetastar {%
southard1987changing}%
\begin{APACrefauthors}%
Southard, D.%
\BCBT {}\ \BBA {} Higgins, T.%
\end{APACrefauthors}%
\unskip\
\newblock
\APACrefYearMonthDay{1987}{}{}.
\newblock
{\BBOQ}\APACrefatitle {Changing movement patterns: Effects of demonstration and
  practice} {Changing movement patterns: Effects of demonstration and
  practice}.{\BBCQ}
\newblock
\APACjournalVolNumPages{Research quarterly for exercise and
  sport}{58}{1}{77--80}.
\PrintBackRefs{\CurrentBib}

\bibitem [\protect \citeauthoryear {%
Stulp%
}{%
Stulp%
}{%
{\protect \APACyear {2012}}%
}]{%
stulp12adaptive}
\APACinsertmetastar {%
stulp12adaptive}%
\begin{APACrefauthors}%
Stulp, F.%
\end{APACrefauthors}%
\unskip\
\newblock
\APACrefYearMonthDay{2012}{}{}.
\newblock
{\BBOQ}\APACrefatitle {Adaptive Exploration for Continual Reinforcement
  Learning} {Adaptive exploration for continual reinforcement learning}.{\BBCQ}
\newblock
\BIn{} \APACrefbtitle {International Conference on Intelligent Robots and
  Systems (IROS)} {International conference on intelligent robots and systems
  (iros)}\ (\BPG~1631-1636).
\PrintBackRefs{\CurrentBib}

\bibitem [\protect \citeauthoryear {%
Stulp%
, Herlant%
, Hoarau%
\BCBL {}\ \BBA {} Raiola%
}{%
Stulp%
\ \protect \BOthers {.}}{%
{\protect \APACyear {2014}}%
}]{%
stulp14simultaneous}
\APACinsertmetastar {%
stulp14simultaneous}%
\begin{APACrefauthors}%
Stulp, F.%
, Herlant, L.%
, Hoarau, A.%
\BCBL {}\ \BBA {} Raiola, G.%
\end{APACrefauthors}%
\unskip\
\newblock
\APACrefYearMonthDay{2014}{}{}.
\newblock
{\BBOQ}\APACrefatitle {Simultaneous On-line Discovery and Improvement of
  Robotic Skill Options} {Simultaneous on-line discovery and improvement of
  robotic skill options}.{\BBCQ}
\newblock
\BIn{} \APACrefbtitle {International Conference on Intelligent Robots and
  Systems (IROS).} {International conference on intelligent robots and systems
  (iros).}
\PrintBackRefs{\CurrentBib}

\bibitem [\protect \citeauthoryear {%
Stulp%
\ \BBA {} Oudeyer%
}{%
Stulp%
\ \BBA {} Oudeyer%
}{%
{\protect \APACyear {2012}}%
{\protect \APACexlab {{\protect \BCnt {1}}}}}]{%
stulp13adaptive}
\APACinsertmetastar {%
stulp13adaptive}%
\begin{APACrefauthors}%
Stulp, F.%
\BCBT {}\ \BBA {} Oudeyer, P\BHBI Y.%
\end{APACrefauthors}%
\unskip\
\newblock
\APACrefYearMonthDay{2012{\protect \BCnt {1}}}{September}{}.
\newblock
{\BBOQ}\APACrefatitle {Adaptive Exploration through Covariance Matrix
  Adaptation Enables Developmental Motor Learning} {Adaptive exploration
  through covariance matrix adaptation enables developmental motor
  learning}.{\BBCQ}
\newblock
\APACjournalVolNumPages{Paladyn. Journal of Behavioral
  Robotics}{3}{3}{128--135}.
\PrintBackRefs{\CurrentBib}

\bibitem [\protect \citeauthoryear {%
Stulp%
\ \BBA {} Oudeyer%
}{%
Stulp%
\ \BBA {} Oudeyer%
}{%
{\protect \APACyear {2012}}%
{\protect \APACexlab {{\protect \BCnt {2}}}}}]{%
stulp12emergent}
\APACinsertmetastar {%
stulp12emergent}%
\begin{APACrefauthors}%
Stulp, F.%
\BCBT {}\ \BBA {} Oudeyer, P\BHBI Y.%
\end{APACrefauthors}%
\unskip\
\newblock
\APACrefYearMonthDay{2012{\protect \BCnt {2}}}{}{}.
\newblock
{\BBOQ}\APACrefatitle {Emergent Proximo-Distal Maturation through Adaptive
  Exploration} {Emergent proximo-distal maturation through adaptive
  exploration}.{\BBCQ}
\newblock
\BIn{} \APACrefbtitle {International Conference on Development and Learning
  (ICDL).} {International conference on development and learning (icdl).}
\newblock
\APACrefnote{{\bf Paper of Excellence Award}}
\PrintBackRefs{\CurrentBib}

\bibitem [\protect \citeauthoryear {%
Stulp%
\ \BBA {} Sigaud%
}{%
Stulp%
\ \BBA {} Sigaud%
}{%
{\protect \APACyear {2012}}%
}]{%
stulp12policy_hal}
\APACinsertmetastar {%
stulp12policy_hal}%
\begin{APACrefauthors}%
Stulp, F.%
\BCBT {}\ \BBA {} Sigaud, O.%
\end{APACrefauthors}%
\unskip\
\newblock
\APACrefYearMonthDay{2012}{}{}.
\newblock
\APACrefbtitle {Policy Improvement Methods: Between Black-Box Optimization and
  Episodic Reinforcement Learning.} {Policy improvement methods: Between
  black-box optimization and episodic reinforcement learning.}
\newblock
\begin{APACrefURL} \url{http://hal.archives-ouvertes.fr/hal-00738463}
  \end{APACrefURL}
\newblock
\APACrefnote{hal-00738463}
\PrintBackRefs{\CurrentBib}

\bibitem [\protect \citeauthoryear {%
Stulp%
\ \BBA {} Sigaud%
}{%
Stulp%
\ \BBA {} Sigaud%
}{%
{\protect \APACyear {2013}}%
}]{%
stulp13robot}
\APACinsertmetastar {%
stulp13robot}%
\begin{APACrefauthors}%
Stulp, F.%
\BCBT {}\ \BBA {} Sigaud, O.%
\end{APACrefauthors}%
\unskip\
\newblock
\APACrefYearMonthDay{2013}{September}{}.
\newblock
{\BBOQ}\APACrefatitle {Robot Skill Learning: From Reinforcement Learning to
  Evolution Strategies} {Robot skill learning: From reinforcement learning to
  evolution strategies}.{\BBCQ}
\newblock
\APACjournalVolNumPages{Paladyn. Journal of Behavioral Robotics}{4}{1}{49--61}.
\PrintBackRefs{\CurrentBib}

\bibitem [\protect \citeauthoryear {%
Thelen%
\ \protect \BOthers {.}}{%
Thelen%
\ \protect \BOthers {.}}{%
{\protect \APACyear {1993}}%
}]{%
thelen1993transition}
\APACinsertmetastar {%
thelen1993transition}%
\begin{APACrefauthors}%
Thelen, E.%
, Corbetta, D.%
, Kamm, K.%
, Spencer, J\BPBI P.%
, Schneider, K.%
\BCBL {}\ \BBA {} Zernicke, R\BPBI F.%
\end{APACrefauthors}%
\unskip\
\newblock
\APACrefYearMonthDay{1993}{}{}.
\newblock
{\BBOQ}\APACrefatitle {The transition to reaching: Mapping intention and
  intrinsic dynamics} {The transition to reaching: Mapping intention and
  intrinsic dynamics}.{\BBCQ}
\newblock
\APACjournalVolNumPages{Child development}{64}{4}{1058--1098}.
\PrintBackRefs{\CurrentBib}

\bibitem [\protect \citeauthoryear {%
Theodorou%
, Buchli%
\BCBL {}\ \BBA {} Schaal%
}{%
Theodorou%
\ \protect \BOthers {.}}{%
{\protect \APACyear {2010}}%
}]{%
theodorou10generalized}
\APACinsertmetastar {%
theodorou10generalized}%
\begin{APACrefauthors}%
Theodorou, E.%
, Buchli, J.%
\BCBL {}\ \BBA {} Schaal, S.%
\end{APACrefauthors}%
\unskip\
\newblock
\APACrefYearMonthDay{2010}{}{}.
\newblock
{\BBOQ}\APACrefatitle {A Generalized Path Integral Control Approach to
  Reinforcement Learning} {A generalized path integral control approach to
  reinforcement learning}.{\BBCQ}
\newblock
\APACjournalVolNumPages{Journal of Machine Learning Research}{11}{}{3137-3181}.
\PrintBackRefs{\CurrentBib}

\bibitem [\protect \citeauthoryear {%
Todorov%
}{%
Todorov%
}{%
{\protect \APACyear {2004}}%
}]{%
todorov04optimality}
\APACinsertmetastar {%
todorov04optimality}%
\begin{APACrefauthors}%
Todorov, E.%
\end{APACrefauthors}%
\unskip\
\newblock
\APACrefYearMonthDay{2004}{}{}.
\newblock
{\BBOQ}\APACrefatitle {Optimality principles in sensorimotor control}
  {Optimality principles in sensorimotor control}.{\BBCQ}
\newblock
\APACjournalVolNumPages{Nature Neuroscience}{7}{9}{907-915}.
\PrintBackRefs{\CurrentBib}

\bibitem [\protect \citeauthoryear {%
Uchibe%
, Asada%
\BCBL {}\ \BBA {} Hosoda%
}{%
Uchibe%
\ \protect \BOthers {.}}{%
{\protect \APACyear {1998}}%
}]{%
Asada98}
\APACinsertmetastar {%
Asada98}%
\begin{APACrefauthors}%
Uchibe, E.%
, Asada, M.%
\BCBL {}\ \BBA {} Hosoda, K.%
\end{APACrefauthors}%
\unskip\
\newblock
\APACrefYearMonthDay{1998}{}{}.
\newblock
{\BBOQ}\APACrefatitle {Environmental Complexity Control for Vision-Based
  Learning Mobile Robot} {Environmental complexity control for vision-based
  learning mobile robot}.{\BBCQ}
\newblock
\BIn{} \APACrefbtitle {roceedings of IEEE International Conference on Robotics
  and Automation} {roceedings of ieee international conference on robotics and
  automation}\ (\BPGS\ 1865--1870).
\newblock
\APACaddressPublisher{}{IEEE Press}.
\PrintBackRefs{\CurrentBib}

\bibitem [\protect \citeauthoryear {%
Vereijken%
, Emmerik%
, Whiting%
\BCBL {}\ \BBA {} Newell%
}{%
Vereijken%
\ \protect \BOthers {.}}{%
{\protect \APACyear {1992}}%
}]{%
vereijken1992free}
\APACinsertmetastar {%
vereijken1992free}%
\begin{APACrefauthors}%
Vereijken, B.%
, Emmerik, R\BPBI E\BPBI v.%
, Whiting, H.%
\BCBL {}\ \BBA {} Newell, K\BPBI M.%
\end{APACrefauthors}%
\unskip\
\newblock
\APACrefYearMonthDay{1992}{}{}.
\newblock
{\BBOQ}\APACrefatitle {Free (z) ing degrees of freedom in skill acquisition}
  {Free (z) ing degrees of freedom in skill acquisition}.{\BBCQ}
\newblock
\APACjournalVolNumPages{Journal of motor behavior}{24}{1}{133--142}.
\PrintBackRefs{\CurrentBib}

\bibitem [\protect \citeauthoryear {%
Vijayakumar%
, D'souza%
\BCBL {}\ \BBA {} Schaal%
}{%
Vijayakumar%
\ \protect \BOthers {.}}{%
{\protect \APACyear {2005}}%
}]{%
vijayakumar2005incremental}
\APACinsertmetastar {%
vijayakumar2005incremental}%
\begin{APACrefauthors}%
Vijayakumar, S.%
, D'souza, A.%
\BCBL {}\ \BBA {} Schaal, S.%
\end{APACrefauthors}%
\unskip\
\newblock
\APACrefYearMonthDay{2005}{}{}.
\newblock
{\BBOQ}\APACrefatitle {Incremental online learning in high dimensions}
  {Incremental online learning in high dimensions}.{\BBCQ}
\newblock
\APACjournalVolNumPages{Neural computation}{17}{12}{2602--2634}.
\PrintBackRefs{\CurrentBib}

\bibitem [\protect \citeauthoryear {%
Westermann%
\ \protect \BOthers {.}}{%
Westermann%
\ \protect \BOthers {.}}{%
{\protect \APACyear {2007}}%
}]{%
westermann2007neuroconstructivism}
\APACinsertmetastar {%
westermann2007neuroconstructivism}%
\begin{APACrefauthors}%
Westermann, G.%
, Mareschal, D.%
, Johnson, M\BPBI H.%
, Sirois, S.%
, Spratling, M\BPBI W.%
\BCBL {}\ \BBA {} Thomas, M\BPBI S.%
\end{APACrefauthors}%
\unskip\
\newblock
\APACrefYearMonthDay{2007}{}{}.
\newblock
{\BBOQ}\APACrefatitle {Neuroconstructivism} {Neuroconstructivism}.{\BBCQ}
\newblock
\APACjournalVolNumPages{Developmental science}{10}{1}{75--83}.
\PrintBackRefs{\CurrentBib}

\end{thebibliography}
}

\end{document}